%% file: main.tex
\documentclass[10pt,twocolumn,letterpaper]{article}

\usepackage{cvpr}              
\usepackage{times}
\usepackage{epsfig}
\usepackage{graphicx}
\usepackage{amsmath}
\usepackage{amssymb}
\usepackage{booktabs}
\usepackage{subcaption}
\usepackage{multirow}
\usepackage{makecell}
\usepackage{booktabs}
\usepackage{comment}
\usepackage{bbding}
\usepackage[accsupp]{axessibility}

\input{preamble}

\definecolor{cvprblue}{rgb}{0.21,0.49,0.74}
\usepackage[pagebackref,breaklinks,colorlinks,citecolor=cvprblue]{hyperref}

\def\paperID{965} 
\def\confName{CVPR}
\def\confYear{2024}

\title{From Isolated Islands to \textit{Pangea}: Unifying Semantic Space for \\Human Action Understanding}

\newcommand*\samethanks[1][\value{footnote}]{\footnotemark[#1]}
\author{
Yong-Lu Li\thanks{The first two authors contribute equally.},~ 
Xiaoqian Wu\samethanks,~
Xinpeng Liu,~
Zehao Wang,~
Yiming Dou,~ 
Yikun Ji,~
Junyi Zhang,\\
Yixing Li,~ 
Jingru Tan,~
Xudong Lu,~ 
Cewu Lu\thanks{Corresponding author.}\\
\tt\small{Shanghai Jiao Tong University}\\
\tt\small{\{yonglu\_li, enlighten, davidwang200099, douyiming,  junyizhang, lyxing0, luxudong2001,} \\
\tt\small{lucewu\}@sjtu.edu.cn, \{xinpengliu0907, jiyikun2002, tanjingru120\}@gmail.com} \\
}

\begin{document}
\maketitle

\begin{abstract}
Action understanding has attracted long-term attention. It can be formed as the mapping from the physical space to the semantic space. Typically, researchers built datasets according to idiosyncratic choices to define classes and push the envelope of benchmarks respectively. Datasets are incompatible with each other like ``\textbf{Isolated Islands}'' due to semantic gaps and various class granularities, \eg, \texttt{do housework} in dataset A and \texttt{wash plate} in dataset B. We argue that we need a more principled semantic space to concentrate the community efforts and use all datasets together to pursue generalizable action learning. To this end, we design a structured action semantic space given verb taxonomy hierarchy and covering massive actions. By aligning the classes of previous datasets to our semantic space, we gather (image/video/skeleton/MoCap) datasets into a unified database in a unified label system, \ie, bridging ``isolated islands'' into a ``\textbf{\textit{Pangea}}''. Accordingly, we propose a novel model mapping from the physical space to semantic space to fully use \textit{Pangea}. In extensive experiments, our new system shows significant superiority, especially in transfer learning. Our code and data will be made public at \href{https://mvig-rhos.com/pangea}{https://mvig-rhos.com/pangea}.
\end{abstract}


\section{Introduction}
Visual action understanding is an important direction in computer vision and matters to various domains~\cite{egger2018automatic,smith2019avid}. 
Generally speaking, it can be formulated as the mapping from the physical space to the semantic space. 
Here, \textit{physical} space indicates the visual patterns (information carrier) and \textit{semantic} space represents the action semantics (class).

\begin{figure}[!t]
    \centering
    \includegraphics[width=\linewidth]{
    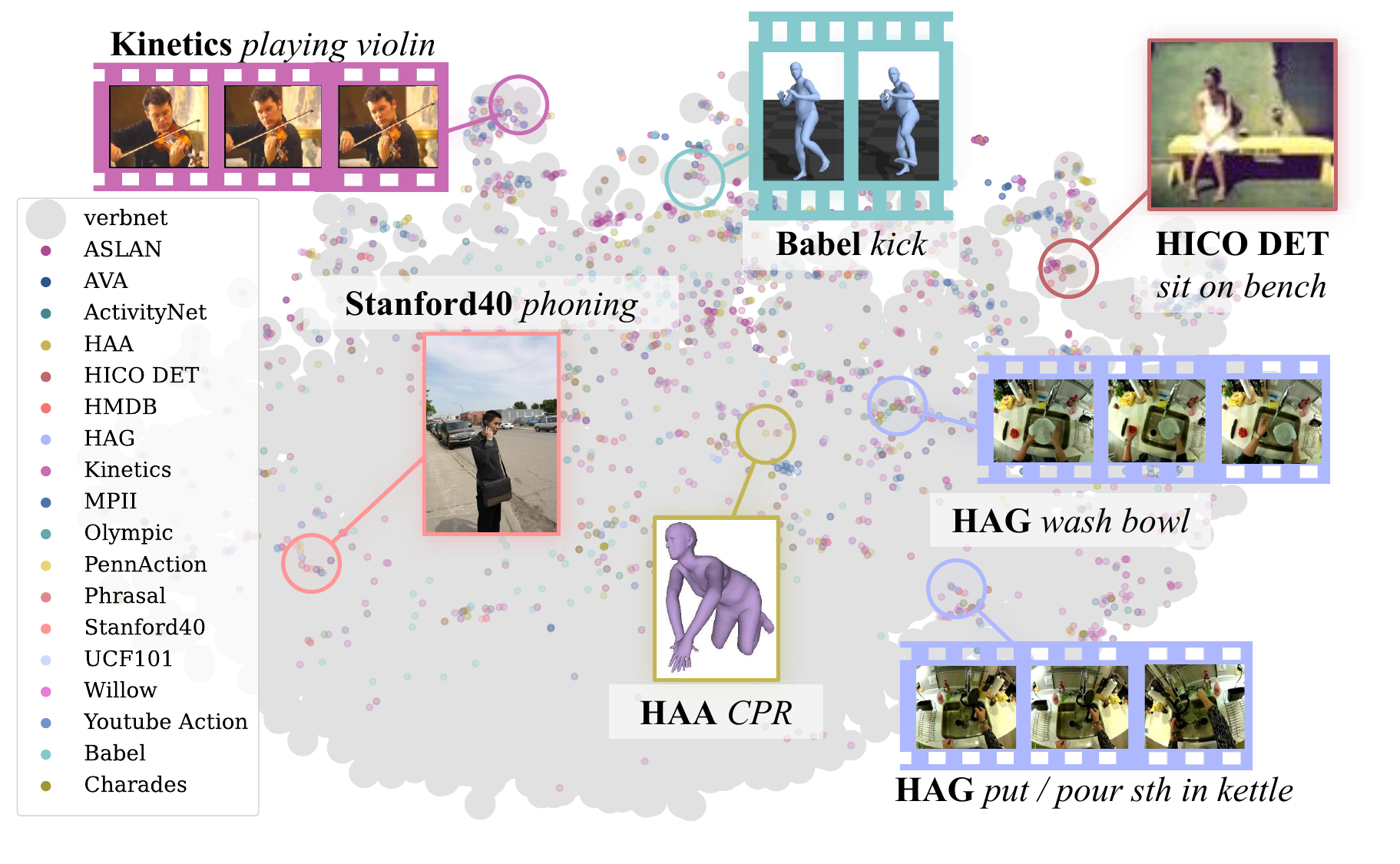
    }
    \vspace{-10px}
    \caption{``Isolated islands''. The semantic gap brings a great challenge to general action understanding.} 
    \label{fig:isolated_islands}
    \vspace{-12px}
\end{figure}

\begin{figure*}[t]
    \centering
    \vspace{-3px}
    \includegraphics[width=0.92\linewidth]{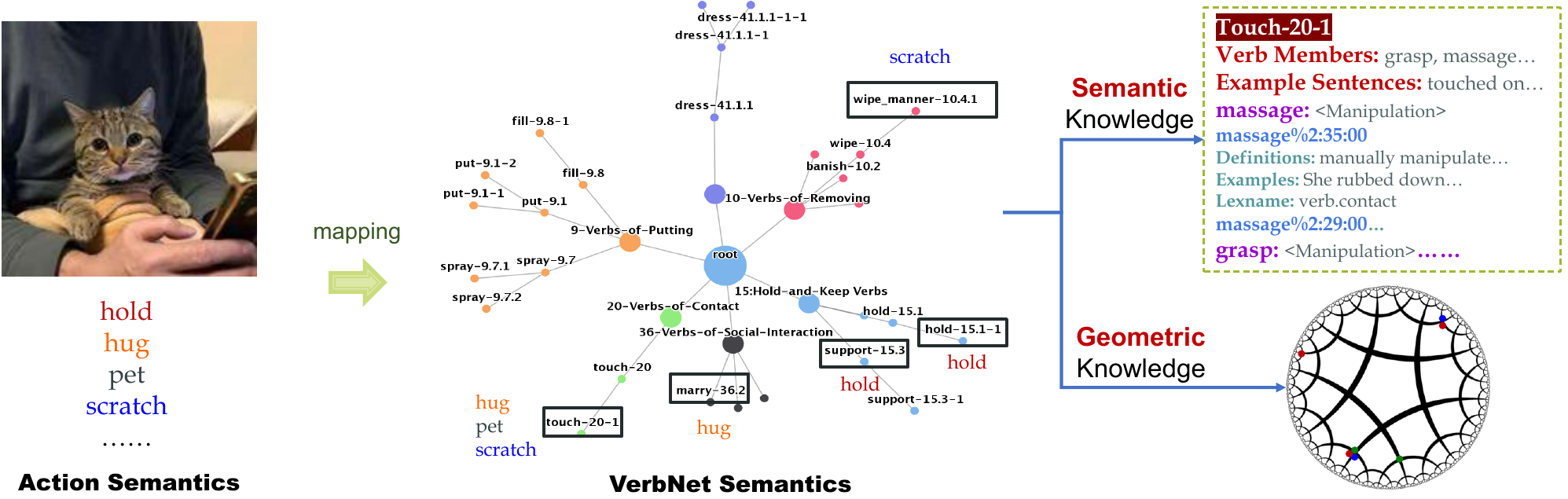} 
    \vspace{-10px}
    \caption{Verb tree. The conventional action semantics (\eg, \texttt{hold, hug}) can be mapped into node semantics (\eg, \texttt{touch-20-1, support-15.3}). The proposed semantic space has abundant semantic and geometric knowledge.}
    \label{fig:verb_node}
    \vspace{-15px}
\end{figure*}

In terms of the physical space, many works were proposed to extract representations from different modalities to capture action cues, such as image~\cite{hakev2}, video~\cite{kinetics-700}, skeleton~\cite{human3}, MoCap~\cite{action2motion}, RGBD~\cite{ntu}, \etc.
However, few efforts have been made to semantic space design. Previous benchmarks~\cite{hico,haa500,kinetics400} are typically designed according to designers' choice and incompatible with each other due to semantic gaps.
They have three main weaknesses:
(1) \textbf{Ambiguity}.
Similar actions may have different class names, \eg, \texttt{clean, wipe, scrub}. Though this may strengthen the diversity in visual-language learning~\cite{clip}, it hinders machines from learning the subtle similarities and differences of actions. 
Besides, the same class may represent different actions, \eg, \texttt{address} means either addressing oneself to something or addressing a conference. This phenomenon brings both generalization possibility and challenge.
(2) Overlooking \textbf{granularity/hierarchy}. 
The datasets are constructed independently, thus typically overlooking granularity, 
\eg, \texttt{do housework} in dataset A and \texttt{clean floor} in dataset B, sometimes even in one dataset.
(3) \textbf{Integration/transfer difficulty}. 
Large models need more data. However, due to the ``isolated islands'', it is hard to integrate datasets and conclude the ``few-shotness'' and ``zero-shotness'' of classes.
We do not know which classes should be enriched or used for transfer learning.

In Fig.~\ref{fig:isolated_islands}, we visualize the class word embeddings~\cite{gao2021simcse} of 18 datasets via t-SNE. Huge semantic gaps exist. Even for the very large Kinetics-700~\cite{kinetics-700}, there are still many classes beyond its coverage.
Here, we first clearly reveal the overlooked ``Isolated Islands (\textbf{I$^2$})'' problem.
It brings semantic gaps and impedes cross-dataset learning. Though CLIP~\cite{clip}-like works alleviate this problem to some extent with the open-vocabulary property, their latent space may be difficult to capture the subtle polysemy, taxonomy, and hierarchy of action semantics. In experiments (Sec.~\ref{sec:experiment}), CLIP trained with simply-mixed datasets performs not well.

Thus, we rethink the action semantic space design and take a step towards a principled semantic space.
We propose a new system to pave a promising way to address the $I^2$ problem. 
Our core idea is to use a \textit{structured} action semantic space to replace the existing hand-crafted ones. 
We build this semantic space according to the linguistic structure knowledge of VerbNet~\cite{verbnet}.
VerbNet is a network linking the syntactic and semantic patterns of verbs. It is a domain-independent tree-structure lexicon and has a clear hierarchy covering most verbs.
We visualize the verb tree in Fig.~\ref{fig:verb_node}.
To maximize the potential of our semantic space, we gather many datasets (image/video/skeleton/MoCap) to build a database and align their classes to our semantic space easily, \ie, linking the ``isolated islands'' into a ``\textit{Pangea}''.
Then, we can use the continuous hyperbolic space together with the semantic-geometric prompt to embed the structured knowledge.

Our space has four-fold superiority:
(1) \textbf{Unambiguous} verb nodes correlating all related verbs, \eg, \texttt{pat, nudge, massage} with similar meaning are shared by the node \texttt{touch-20-1}.
(2) \textbf{Rich knowledge}. Besides the thematic role, syntactic, semantic description, and selectional preferences of verbs, VerbNet has mappings to other knowledge bases (WordNet~\cite{wordnet}, PropBank~\cite{propbank}, FrameNet~\cite{framenet}).
We can conveniently adopt Large Language Models~\cite{gpt3} to extract meaningful language representations to advance learning.
(3) \textbf{Hierarchy} to represent actions from abstract to specific granularity, \eg, \texttt{sports, ball sports, basketball, dunk}.
(4) \textbf{Extensive coverage}. It contains about 5,800 verbs.
In Fig.~\ref{fig:isolated_islands}, 
our space not only covers all datasets but also spans the semantics a lot. 

To fully use \textit{Pangea}, we propose a compact mapping system to conduct action understanding, which effectively maps multi-modal physical patterns to the structured semantic space.
In experiments, our method armed with \textit{Pangea} demonstrates representative and transfer ability. On multi-modal benchmarks, it brings decent improvements. 

Our contributions are: 
1) We propose a structured semantic space to bridge the ``isolated islands''.
2) We build the \textit{Pangea} database gathering 28 multi-modal datasets.
3) A physical-to-semantic mapping model is proposed given \textit{Pangea} and shows significant transfer ability.

\section{Related Work}
\label{sec:related_work}
Action Understanding has achieved progress recently. There are mainly image~\cite{hico,vcoco,yao2011human}, video~\cite{AVA,kinetics400,soomro2012ucf101,caba2015activitynet}, skeleton~\cite{liu2019ntu}, and 3D body~\cite{BABEL:CVPR:2021} datasets.
The common tasks are action recognition and temporal/spatial localization/detection. Early benchmarks focus on classifying an image or a short video into classes~\cite{soomro2012ucf101,hico,yao2011human}. Recently, benchmarks that require both accurate recognition and active subject detection are emerging~\cite{AVA,hicodet,vcoco,liu2022interactiveness,wu2022mining,liu2022highlighting,idn}.
Moreover, few/zero-shot action learning~\cite{chen2021elaborative} also attracts attention.  
Many methods have been proposed to push this direction forward.
For image tasks, 2D CNN is the dominant architecture, while knowledge like part state~\cite{partstate,pastanet}, 2D/3D human~\cite{djrn,tincvpr}, and language prior~\cite{functional,analogy,vcl,hakev2} is used too. 
For video tasks, 2D-CNN~\cite{yue2015beyond,donahue2015long,lin2019tsm}, two-stream network~\cite{feichtenhofer2016convolutional,simonyan2014two}, and 3D-CNN~\cite{kinetics-700,feichtenhofer2019slowfast,dio} are the major architectures adopted. 
For skeleton tasks, both GCN~\cite{yan2018spatial,li2019actional,liu2020disentangling} and 2D-CNN~\cite{choutas2018potion,yan2019pa3d} are widely used.
Recently, with the success of Transformer~\cite{Vaswani2017AttentionIA,Symbol-LLM}, besides directly importing it into action detection~\cite{detr,qpic}, visual-language contrastive learning~\cite{clip} has changed this direction a lot.

In terms of action semantic space, most datasets~\cite{hico,vcoco,AVA,ASLAN,hmdb} overlook action hierarchy.
While some works consider hierarchy~\cite{caba2015activitynet,shao2020finegym,hyperbole}.
For example, ActivityNet~\cite{caba2015activitynet} defines 200+ action classes belonging to 7 high-level classes (\eg, \texttt{personal care, household}) based on activity scenarios;
FineGym~\cite{shao2020finegym} organizes hierarchical actions from gymnasium videos;
VerSe~\cite{verse}
augments COCO~\cite{coco} and TUHOI~\cite{tuhoi} with verb sense labels to provide finer-grained action semantics on 3.5 K images.
However, they are scale/class/domain-limited and built with manually-picked classes. 
Instead, we choose to cover the hierarchy based on well-defined linguistic works such as VerbNet~\cite{verbnet}, WordNet~\cite{wordnet}, FrameNet~\cite{framenet}, \etc. 

\section{Preliminary}
\label{sec:preliminary}
In this section, we first introduce the preliminaries of the physical and semantic space.

{\bf Multi-Modal Physical Space}.
Here, we adopt two modalities for physical space $P$: 2D and 3D.
For 2D, we adopt CNN or Transformer (\eg, ResNet~\cite{resnet}, CLIP~\cite{clip}) to extract representation from image/video. 
For 3D, we use the widely-used model SMPL~\cite{SMPL} to embed 3D humans. 

{\bf Structured Semantic Space}.
Intuitively, the ambiguity of objects is relatively smaller, thus objects/nouns are easier to label. 
Things are different for actions/verbs which are more ambiguous.
Previous works typically design semantic space manually and optionally. 
Instead, we build the structured semantic space $S$ via the hierarchical verb tree from VerbNet~\cite{verbnet} (Fig.~\ref{fig:verb_node}).
Here, we define the \textit{nodes} as the \textit{classes} of our semantic space. 
Compared with conventional design~\cite{caba2015activitynet}, our space has elegant characteristics:
\textbf{(1)} Due to the lack of a unified naming standard, classes of previous datasets have ambiguity. For example, different datasets may have \texttt{feast}, \texttt{eating}, and \texttt{dining} respectively, where a common semantic is shared. 
Instead, in our $S$, actions with \textbf{shared} meanings are connected with their \textit{common} nodes.
\textbf{(2)} Each node is equipped with \textbf{abundant knowledge}. 
In Fig.~\ref{fig:verb_node}, \texttt{touch-20-1} node is explained by: 
    a) Verb members, \eg, \texttt{grasp};
    b) Example sentences as instantiations of the node semantics;
        c) Each verb member is explained via connections with other lexical resources (\eg, WordNet~\cite{wordnet}, FrameNet~\cite{framenet}). In Fig.~\ref{fig:verb_node}, the verb \texttt{massage} is explained by its frame in FrameNet~\cite{framenet} (\texttt{manipulation}) and the corresponding items in WordNet~\cite{wordnet} (\texttt{massage\%2:35:00}, \texttt{massage\%2:29:00}).
\textbf{(3)} \textbf{Hierarchy} reveals semantic connections between nodes and provides structured knowledge. The nodes are numbered according to shared semantics and syntax. Nodes sharing a high-level number (9-109) have semantic relations~\cite{verbnet}, \eg, \texttt{banish-10.2} and \texttt{wipe-10.4} share a parent node as they are all about \texttt{removing}. 
Though some works~\cite{caba2015activitynet,shao2020finegym} consider hierarchy too, they are either of limited coverage or defined empirically according to scenes. 
Instead, our verb semantics are more explicit.
\textbf{(4)} Our $S$ covers \textbf{5,800+} verbs which is broader than previous works.

\section{Constructing \textit{Pangea}}
\label{sec:database}
{\bf Data Curation}.
With the structured $S$, we collect data with diverse modalities, formats, and granularities, and adapt them into a unified form.  
Our database \textit{Pangea} contains a large range of data including image, video, and skeleton/MoCap. 
We process and formulate them as follows:

\begin{figure}[t]
    \centering
    \includegraphics[width=0.95\linewidth]{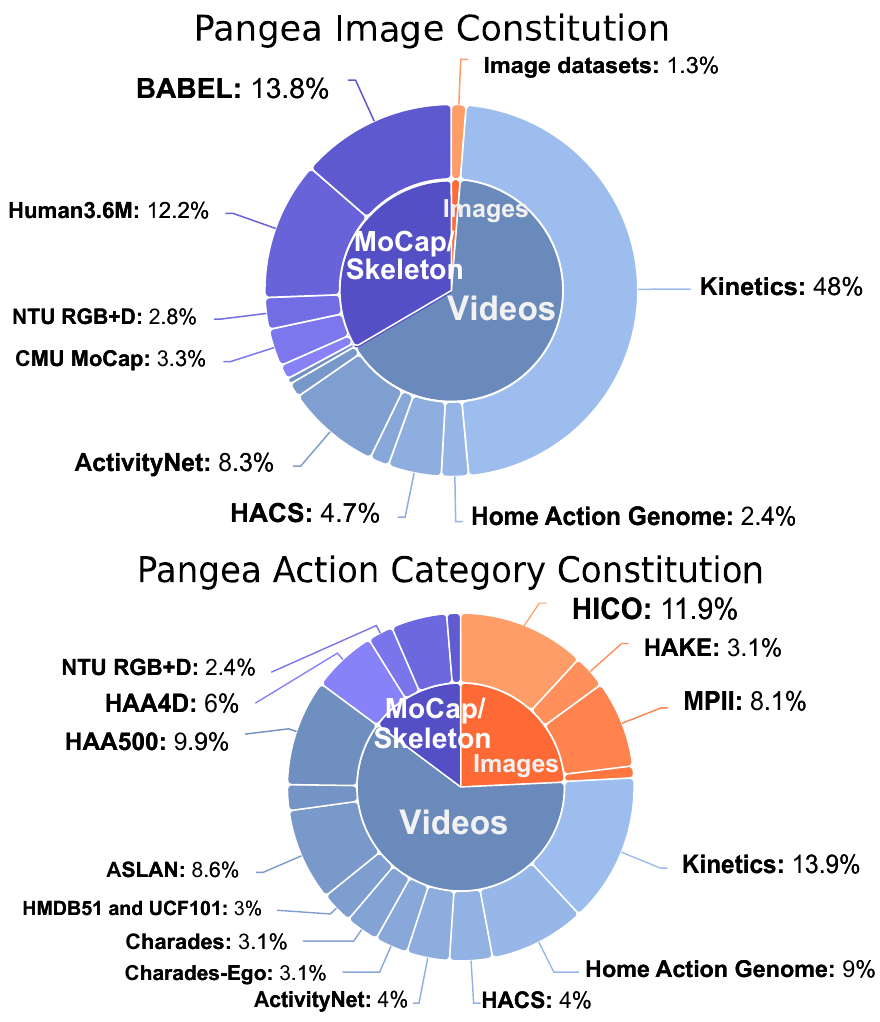} 
    \vspace{-10px}
    \caption{Gathered datasets in \textit{Pangea}. 
    }
    \vspace{-15px}
    \label{fig:dataset_list}
\end{figure}
    
1) \textbf{Semantic Consistency.} 
The class definitions of datasets are various, but they can be mapped to our semantic space with the fewest semantic damages. The mapping is completed via manual annotation with the help of word embedding~\cite{clip} distances and OpenAI GPT-3.5.  
Manual annotation is the most accurate and most expensive, while word embedding comparison is the least.
Thus, we adopt a hybrid method: potential class-node mapping is first filtered out roughly by comparing word embedding, then selected via GPT-3.5 prompting, and finally checked by human annotators.
As more and more classes are aligned and covered, the process would be faster and faster with synonyms checking.
As shown in  Fig.~\ref{fig:isolated_islands}, our semantic space covers a broad range of semantics, verifying this mapping.

2) \textbf{Temporal Consistency.} 
Some videos~\cite{kinetics-700} only have sparse labels for a whole clip instead of each frame. 
For these sparse datasets, we sample the clip with 3 FPS and give frames the label of their belonged clip.
We provide both frame- and clip-level labels. 
    
3) \textbf{Spatial Consistency.} 
There are both instance (boxes)~\cite{hicodet} and image~\cite{hico} level labels. 
We merge the instance labels of each image/frame into image/frame-level labels.
For demands of instance-level training, we can use the original instance labels~\cite{AVA,hicodet} and detectors~\cite{faster,detr} to get instance boxes even masks~\cite{sam} for future annotation.
    
4) \textbf{3D Format Consistency.} 
3D datasets typically have various formats, \eg, SMPL~\cite{SMPL} has $24$ keypoints while CMU MoCap~\cite{action2motion} has $31$ keypoints. To keep consistency, we transform all of them into SMPL via a fitting procedure. 

\begin{figure}[t]
    \centering
    \vspace{-5px}
    \includegraphics[width=\linewidth]{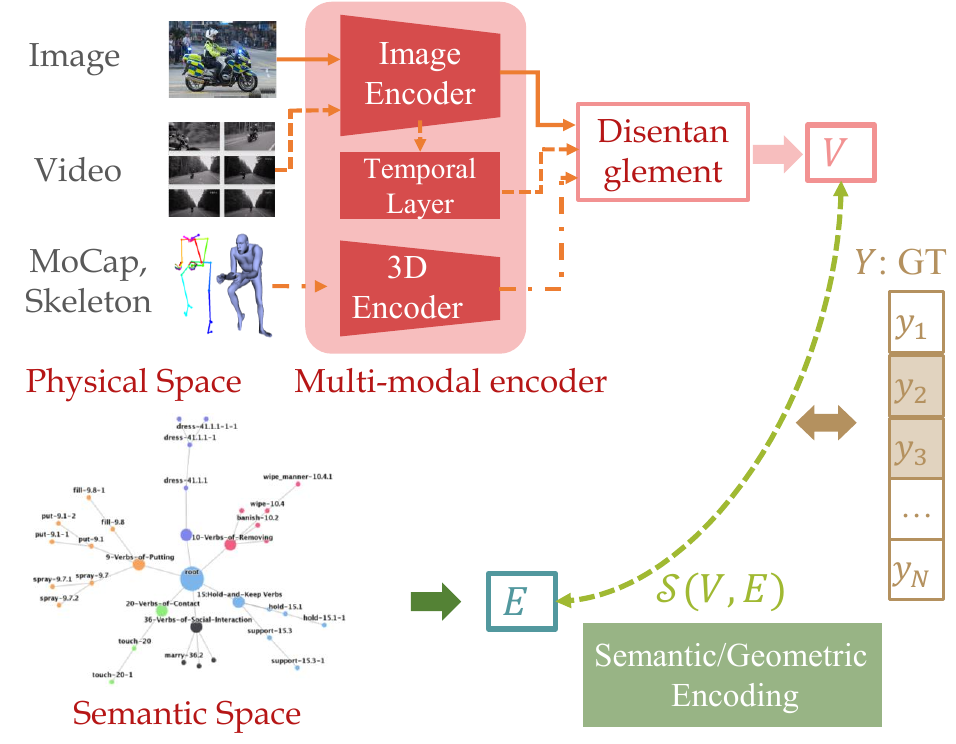} 
    \vspace{-10px}
    \caption{P2S mapping. Given a sample, we obtain its representation $V$ via encoders. $V$ is then aligned with node representations $G$ under the supervision of $Y$. $v_{raw}$ to $V$ is omitted for clarity.} 
    \label{fig:overview}
    \vspace{-15px}
\end{figure}
    
5) \textbf{2D-3D Consistency.} 
Image/video datasets mostly contain only 2D labels without GT 3D humans. 
Aside from the GT 3D humans from 3D datasets~\cite{BABEL:CVPR:2021}, we recover 3D humans from 2D data as pseudo 3D labels via ROMP~\cite{romp} and EFT~\cite{joo}. 
We use both GT and pseudo 3D humans in 3D action recognition.
Though the reconstruction is sometimes noisy, we use the pseudo 3D humans as noisy data \textit{augmentation} to supplement 2D learning. 
In tests, we find that 2D and 3D learning are complementary. 

{\bf Analysis}. 
With the large data collection and unified semantic space, we build \textit{Pangea} as shown in Fig.~\ref{fig:dataset_list}. 
It contains 19.5 M images, 1.1 M videos, 
and 840 K 3D humans over 28 datasets, with coverage of 4 K+ action classes of original datasets.  
\textit{Pangea} covers the semantics of 513 verb nodes over all the 898 nodes of VerbNet~\cite{verbnet} and includes 290 leave nodes carrying fine-grained semantics.

\section{Methodology}
\label{sec:method}

\subsection{Overview of P2S Mapping}
First, we introduce the Physical-to-Semantic Space (P2S) mapping (Fig.~\ref{fig:overview}).
We aim to propose a multi-modal, concise, and 
practical model as the baseline to inspire future work.
Given a sample of the physical space $P$, we obtain its representation $V$ via different encoders according to its modality.
For images, we use a CNN/Transformer-based image encoder. 
For videos, we first input them to the image encoder for frame encoding and then use a temporal layer for temporal encoding.
For SMPLs, we covert them into point clouds and use a PointNet++\cite{PointNet++} as the encoder.

In the semantic space $S$, we define $N$ target verb nodes. For each node, two types of information are provided by VerbNet~\cite{verbnet}: 
1) \textbf{semantic} one to describe its meaning, \eg, example sentences, WordNet definitions; 
2) \textbf{geometric} one to locate it in the tree and reveal its connection with the other nodes.
The semantic and geometric information can be encoded via the verb node representation $E=\{e_i\}^N_{i=1}$ (detailed in Sec.~\ref{sec:encoding}).
The ground-truth (GT) label for the sample is $Y=\{y_i|y_i \in \ \{0, 1\}\}_{i=1}^N$.
P2S mapping is a multi-label classification, where a physical sample is mapped to multi-node of the semantic space (\textbf{one-to-many} mapping). 
The similarity $\mathcal{S}(V, E)$ between $V$ and $E$ is bound with the GT label $Y$, and the loss function is derived in Sec.~\ref{sec:encoding}.
In Sec.~\ref{sec:disentangle}, we discuss how to facilitate such one-to-many mapping with semantic disentangle and augmentation. We summarize the training and inference in Sec.~\ref{sec:train_infer}.

\subsection{Semantic Disentanglement and Augmentation}
\label{sec:disentangle}
A person typically performs multi-action simultaneously, \eg, standing while eating. Such entanglement of multi-action semantics increases the annotation and learning difficulty. It is a challenge to annotate all the ongoing actions of a person in previous datasets due to the limited coverage and ambiguity of their classes. 
Besides, as \textit{Pangea} has a broader semantic space, after the \textit{action}$\rightarrow$\textit{node} mapping in Sec.~\ref{sec:database}, we face a \textbf{partial-label} learning problem.
Moreover, in the mapping, inevitably, some labels are early filtered out and a few of them should have been annotated as True. 
Also, errors of omission may exist within the labels because of annotators' bias.
Thus, each sample theoretically has a partial annotation $Y=\{y_i|y_i=1,0,\emptyset\}_{i=1}^N$, where $1,0$ are certain positive/negative labels, and $\emptyset$ is uncertain.
Though it is nearly impossible to supplement the labels of all $N$ verb nodes in \textit{Pangea} for all samples (images/videos/MoCap), we can conduct flexible weakly-supervised learning with partly-labeled data with representation disentanglement.

\begin{figure}[t]
    \centering
    \vspace{-5px}
    \includegraphics[width=\linewidth]{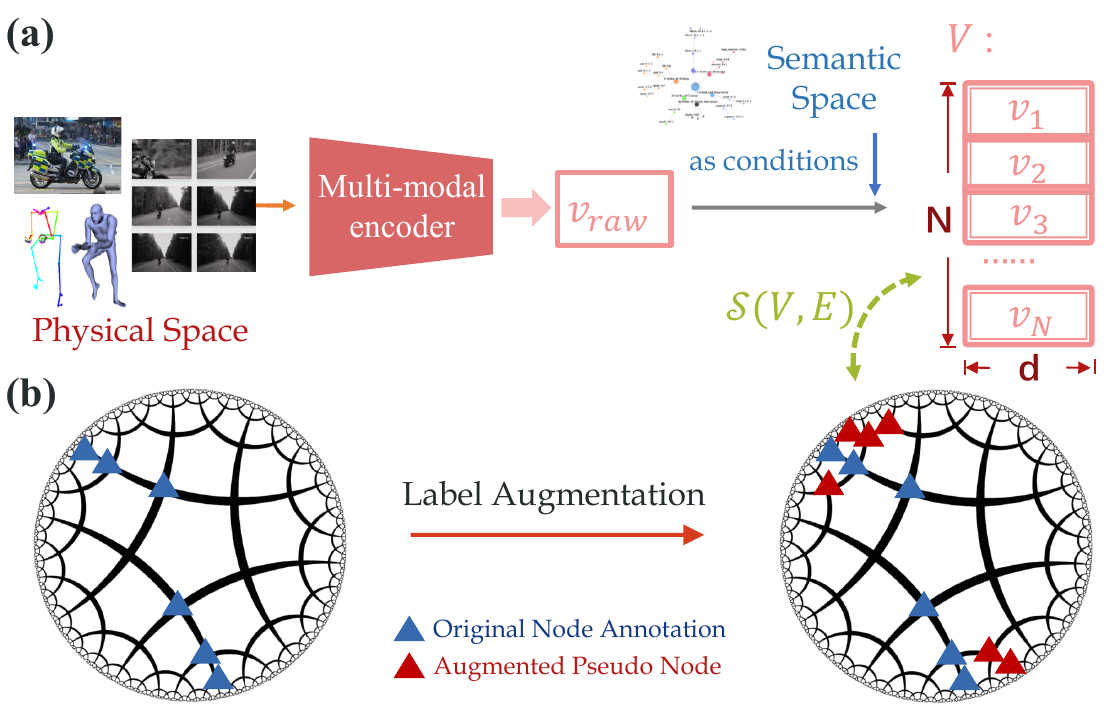}
    \vspace{-10px}
    \caption{Semantic disentanglement and augmentation. (a) Entangled physical representation $v_{raw}$ is mapped into node-specific $V=\{v_i\}_{i=1}^N \in \mathbf{R}^{N \times d}$ ($i$: node index as \textit{conditions}).
    (b) Getting pseudo node labels via language priors and structure knowledge.
    } 
    \label{fig:dis+aug}
    \vspace{-12px}
\end{figure}

To facilitate the one-to-many P2S mapping and address the partial-label learning problem, we propose disentangling a physical representation into \textbf{node-specific} representations.
We use $v_{raw}$ as the entangled physical feature.
Thanks to our unambiguous verb node definition, we disentangle the input $v_{raw}$ into $N$ representations supervised by $N$ verb nodes respectively. 
Thus, the gradients of verb nodes (True/False labeled clearly) were disentangled during training from the uncertain ones.
As is illustrated in Fig.~\ref{fig:dis+aug}, a model is trained to transform the entangled physical representation $v_{raw} \in \mathbf{R}^d$ ($d$: dimension) into node-specific representation $V=\{v_i\}_{i=1}^N \in \mathbf{R}^{N \times d}$ ($i$: verb node index) as conditions.
To get $V=\{v_i\}_{i=1}^N \in \mathbf{R}^{N \times d}$, we first define the verb node-specific disentangling mapping function $f_{i}$ for the $i$-th node, which is in practice a learnable MLP. 
Then, given $v_{raw}$, $f_{i}(\cdot)$ transforms it into $v_i= f_{i}(v_{raw})$.
To apply P2S mapping with this disentangled physical representation, the similarity between $V$ and $G$ is measured as 
\begin{equation}
\mathcal{S}(V, E) = \{\mathcal{S}(v_i, e_i)\}_{i=1}^N,
\label{eq:w_disentangle}
\end{equation}
If not disentangled, it goes like 
\begin{equation}
\mathcal{S}(v_{raw}, E) = \{\mathcal{S}(v_{raw}, e_i)\}_{i=1}^N,
\label{eq:wo_disentangle}
\end{equation}
\ie, physical representation is shared by all verb nodes.

Aside from disentanglement, the partial-label learning problem can be alleviated by augmenting the GT label $Y$. 
As the verb nodes in our structured semantic space have clear semantic and geometric relations in a tree, we propose a label augmentation method to generate pseudo node labels for missing ones via language priors and structure knowledge.
For more details, please refer to Suppl. Sec.~\textcolor{red}{3}.

\subsection{Verb Node Encoding and Alignment}
\label{sec:encoding} 

\subsubsection{Semantic Encoding}
\label{sec:s_align}
Next, we discuss how to use text representation to encode semantic information of nodes into $E =\{e_i\}_{i=1}^N \in \mathbf{R}^{N \times d}$.
As mentioned in Sec.~\ref{sec:preliminary}, a verb node is composed of several actions with shared meanings.
The node semantic information includes: 
1) verb members; 
2) example sentences; 
3) WordNet~\cite{wordnet} definition and FrameNet~\cite{framenet} mapping for each verb member.
Following CLIP~\cite{clip} text encoder, we get $E$ via inputting these texts into a Transformer encoder. 

Different from CLIP~\cite{clip} where the text is short (up to 77 tokenized symbols, or equally 30 words approximately), our node description can be longer when the node contains many verb members.
It is inefficient, unstable, and memory-costly to input such long text into the encoder directly.
Thus, we sample key texts clarifying the node semantics better. We use TextRank~\cite{PyTextRank} to extract keywords and then take the \textit{summarized} text as the text encoder input.

\subsubsection{Geometric Encoding}
\label{sec:g_align}
\begin{figure*}[!t]
    \centering
    \vspace{-5px}
    \includegraphics[width=0.9\linewidth]{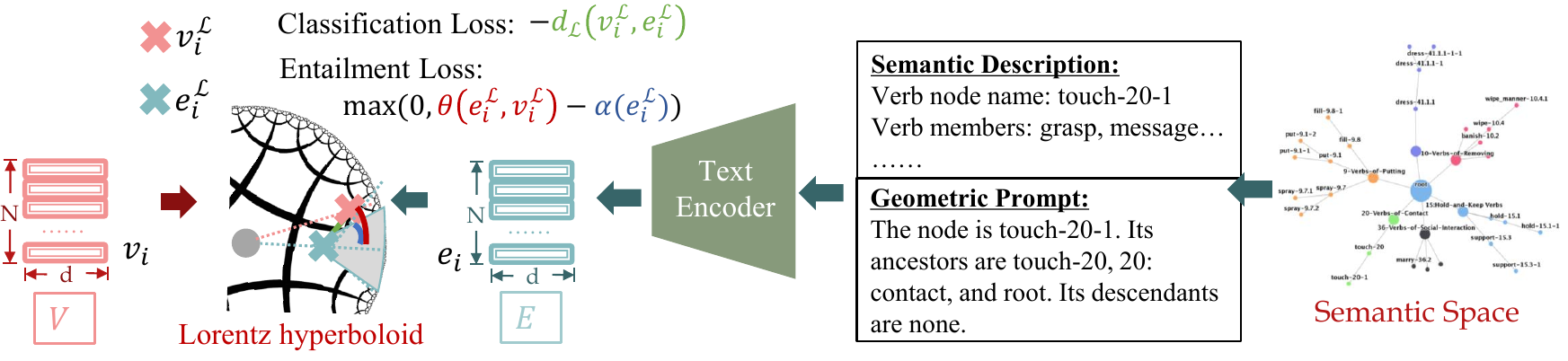} 
    \vspace{-10px}
    \caption{Verb node encoding and $V-E$ alignment. 
    1) The \textit{right} part: encoding semantic and geometric information via a text encoder.
    2) The \textit{left} part: $V-E$ alignment in a Lorentz hyperboloid with two training objectives: classification loss and entailment loss.
    } 
    \label{fig:geo_align}
    \vspace{-9px}
\end{figure*}

Next, we encode the geometric information into $E =\{e_i\}_{i=1}^N \in \mathbf{R}^{N \times d}$. 
To encode the hierarchy, parent-child relation, verb tree depth, \etc, we use hyperbolic representations~\cite{desai2023hyperbolic} of the physical representation $V$ and verb node representation $E$.
Besides, to utilize the representative ability of language models~\cite{bert,clip}, we also propose a geometric prompt strategy to strengthen the training.
Fig.~\ref{fig:geo_align} is the overview of the encoding and $V-E$ alignment processes.

{\textbf{Geometric Prompt.}}
A direct way is to use language descriptions as prompts, \eg, the node \texttt{touch-20-1} is described as: ``The node is \texttt{touch-20-1}. Its ancestors are \texttt{touch-20, 20}: contact, and \texttt{root}. Its descendants are none.'' 
We use a text encoder to encode these prompts. 
In practice, we use text concatenation to integrate the geometric descriptions and those semantic descriptions introduced in Sec.~\ref{sec:s_align}.
We concatenate these sentences and input them together into one Transformer encoder to get $E$.
        
{\textbf{Hyperbolic Representation}}
The proposed semantic space is hierarchical, revealing connections between nodes and providing structured knowledge (Sec.~\ref{sec:preliminary}). 
The text description of nodes implicitly reveals the hierarchy. 
For example, a node with a text description ``The node is \texttt{put-9.1.1}. Its ancestors are \texttt{put-9.1, 9}: putting, and root. Its descendants are none. Its verbs are: apply, insert, install ...'' would be closer to \texttt{put-9.1} (more generic concepts) and \texttt{put-9.1.2} (neighbor).
Besides, P2S is a verb node multi-label classification. Thus, one physical representation can be aligned with both generic concepts which are closer to the \textit{root} node of the hierarchy (\eg, \texttt{10:removing}), and specific concepts which are closer to the \textit{leaf} node (\eg, \texttt{banish-10.2}, \texttt{wipe-10.4}).
Thus, Euclidean space is not suitable for our task, which applies the same distance metric to all embedded points. 

Here, we leverage the hyperbolic representation~\cite{nickel2017poincare} which can capture hierarchy to embed $V$ and $E$.
Specifically, we adopt a Lorentz model of hyperbolic geometry~\cite{desai2023hyperbolic}. Thus, similar to~\cite{desai2023hyperbolic}, the semantic hierarchy emerges in the representation space. We can thus align each disentangled physical representation to its corresponding multiple node representations.
For a detailed formulation of the Lorentz model, please refer to Suppl. Sec.~\textcolor{red}{3}.

There are two objectives in the alignment: classification loss and entailment loss. Fig.~\ref{fig:geo_align} illustrates the calculation.

{\textbf{Classification Loss.}}
We have the disentangled physical representation $V=\{v_i\}_{i=1}^N$, node representation $E=\{e_i\}_{i=1}^N$, and GT label $Y=\{y_i| y_i \in \{0,1\}\}_{i=1}^N$. 
For each $i$, $v_i$ and $e_i$ are first mapped into $v_i^{\mathcal{L}}$ and $e_i^{\mathcal{L}}$ in the Lorentz hyperboloid via \textit{exponential map}. The similarity $\mathcal{S}(v_i, e_i)$ is measured via the negative of Lorentzian distance $d_{\mathcal{L}}(\cdot, \cdot)$ between $v_i^{\mathcal{L}}$ and $e_i^{\mathcal{L}}$. 
Thus, the classification loss is:
\begin{equation}
\mathcal{L}_{cls} = \mathcal{L}_{BCE}(\{Sigmoid(\gamma \cdot -d_{\mathcal{L}}(v_i^{\mathcal{L}}, e_i^{\mathcal{L}}))\}_{i=1}^N, Y),
\label{eq:cls_loss}
\end{equation}
where $\gamma$ is a scaling factor.
For multi-label classification, the output is processed by a Sigmoid function and bound with Binary Cross Entropy (BCE) loss. 

{\textbf{Entailment Loss.}}
In addition to the classification loss, an entailment loss is applied to enforce partial order relationships between the node representation $e_i^{\mathcal{L}}$ and physical representation $v_i^{\mathcal{L}}$.
If $y_i=1$, the physical representation $v_i^{\mathcal{L}}$ should lie inside the entailment cone~\cite{ganea2018hyperbolic} of the node representation $e_i^{\mathcal{L}}$.
As is illustrated in Fig.~\ref{fig:geo_align}, it is measured by comparing the exterior angle $\theta(e_i^{\mathcal{L}}, v_i^{\mathcal{L}})$ and the half-aperture $\alpha(e_i^{\mathcal{L}})$.
Thus, the entailment loss is :
\begin{equation}
\mathcal{L}_{ent} = \frac{1}{sum(Y)}\sum\limits_{i: y_i=1} max(0, \theta(e_i^{\mathcal{L}}, v_i^{\mathcal{L}}) - \alpha(e_i^{\mathcal{L}})).
\end{equation}
The loss functions as a further constraint besides the classification loss.

\subsection{Training and Inference}
\label{sec:train_infer}
In training, the total loss $\mathcal{L}_{total} = \mathcal{L}_{cls} + \omega \mathcal{L}_{ent}$, where $\omega$ balances the loss weight (here $\omega=0.01$).
In inference, P2S outputs probabilities of verb nodes $\mathcal{S}_{node}=\{Sigmoid(\gamma \cdot -d_{\mathcal{L}}(v_i^{\mathcal{L}}, e_i^{\mathcal{L}}))\}_{i=1}^N$ from Eq.~\ref{eq:cls_loss}.
We evaluate node classification with $\mathcal{S}_{node}$ on \textit{Pangea} test set (Sec.~\ref{sec:exp_result}).
For transfer learning, we pretrain P2S on \textit{Pangea} and finetune it on downstream datasets.
To get the action class score $\mathcal{S}_{act}$ of the downstream dataset, we fix the node prediction and use a small learnable MLP to transform $\mathcal{S}_{node}$ to $\mathcal{S}_{act}$.

\begin{table}[t]
    \centering
    \resizebox{0.55\linewidth}{!}{
    \begin{tabular}{l | c c c }
        Method  & Full & Rare & Non-Rare \\
        \toprule[2pt]
        CLIP    & 28.25  &  16.90  &37.87 \\ 
        \hline
        P2S   & 34.01   & 21.37  & 44.72  \\ 
        P2S-aug & \textbf{34.25} & \textbf{21.56} & \textbf{45.00} \\
      \end{tabular}}
      \vspace{-10px}
      \caption{Verb node classification results on \textit{Pangea} test set.} 
    \label{tab:node_benchmark}
    \vspace{-18px}
\end{table}

\section{Experiment}
\label{sec:experiment}

\begin{figure*}[t]
\centering
\vspace{-3px}
\begin{minipage}{0.23\textwidth}
    \centering
    \resizebox{0.8\linewidth}{!}{
    \begin{tabular}{l | c }
        Method  &  mAP\\
        \toprule[2pt]
        R*CNN~\cite{gkioxari2015contextual} & 	28.50 \\
        Mallya \etal ~\cite{mallya2016learning} & 36.10 \\
        Pairwise~\cite{Fang2018Pairwise} & 39.90 \\
        RelViT~\cite{ma2022relvit} & 40.12 \\ 
        \toprule[2pt]
        CLIP~\cite{clip}  &  46.35 \\ 
        \hline
        \makecell[l]{CLIP-\textit{Pangea}}    & 45.09  \\ 
        P2S    & \textbf{47.74}  \\  
      \end{tabular}}
      \vspace{-10px}
        \captionof{table}{Results on the image benchmark HICO~\cite{hico}.} 
        \label{tab:hico}
\end{minipage}
\hfill
\begin{minipage}{0.26\textwidth}
    \centering
    \resizebox{\linewidth}{!}{
    \begin{tabular}{l | c}
        Method  &  Top-1 Accuracy (\%) \\
        \toprule[2pt]
        I3D~\cite{I3D} & 33.53	\\
        TPN~\cite{TPN} & 50.53\\
        TSN~\cite{TSN} & 55.33\\
        EVL~\cite{lin2022frozen} & 76.40\\
        \toprule[2pt]
        CLIP~\cite{clip}  &  66.33 \\
        \hline
        CLIP~\cite{clip}-\textit{Pangea} & 68.27 \\
        P2S & 71.40  \\ 
        P2S + EVL~\cite{lin2022frozen} & \textbf{80.87}\\ 
      \end{tabular}}
      \vspace{-10px}
        \captionof{table}{Results on the video benchmark HAA~\cite{haa500}.} 
        \label{tab:haa}
\end{minipage}
\hfill
\begin{minipage}{0.3\textwidth}
    \centering
    \resizebox{\linewidth}{!}{
    \begin{tabular}{l | c c c c }
        Method  &  Top-1 Accuracy (\%)(all)\\
        \toprule[2pt]
        TSN~\cite{TSN} & 69.40 \\
        RGB-I3D~\cite{I3D}  & 74.30 \\
        Two-stream I3D~\cite{I3D} & 80.90 \\
        EVL~\cite{lin2022frozen} & 83.68\\
        \toprule[2pt]
        CLIP~\cite{clip}  & 67.47 \\
        \hline
        CLIP~\cite{clip}-\textit{Pangea}  &67.69 \\
        P2S   & 68.37 \\ 
        P2S + EVL~\cite{lin2022frozen} &\textbf{85.09}\\
      \end{tabular}}
      \vspace{-10px}
        \captionof{table}{Results on the video benchmark HMDB51~\cite{hmdb}.} 
        \label{tab:hmdb}
\end{minipage}
\vspace{-10px}
\end{figure*}

\begin{figure*}[t]
\centering
\begin{minipage}{0.32\textwidth}
    \centering
    \resizebox{\linewidth}{!}{
    \begin{tabular}{l | c c}
        Method  &  Top-1 Acc (\%) & Top-5 Acc (\%)\\
        \toprule[2pt]
        TSN~\cite{TSN} & 73.90	& 91.10 \\
        VideoMAE~\cite{tong2022videomae} & 87.40 & 97.60 \\
        EVL~\cite{lin2022frozen}    & 87.64	& 97.71  \\ 
        \toprule[2pt]
        CLIP~\cite{clip}  &  72.82 & 91.68 \\ 
        \hline
        CLIP~\cite{clip}-\textit{Pangea}   & 70.45 & 89.14 \\ 
        P2S     & 73.80 & 92.01 \\ 
        P2S + EVL~\cite{lin2022frozen} & \textbf{90.22}& \textbf{98.26}\\
      \end{tabular}}
      \vspace{-10px}
        \captionof{table}{Results on the video benchmark Kinetics-400~\cite{kinetics400}.}
        \label{tab:kinetics}
\end{minipage}
\hfill
\begin{minipage}{0.35\textwidth}
    \centering
    \resizebox{\linewidth}{!}{
    \begin{tabular}{l|c c}
        Methods             & Top-1\% & Top-1-norm\% \\
        \toprule[2pt]
        2s-AGCN~\cite{BABEL:CVPR:2021}                & 40.00 & 16.00 \\
        PointNet++~\cite{PointNet++}                  & 42.26 & 24.73\\
        CLIP~\cite{clip}                              & 32.42 & 9.84\\
        \hline
        PointNet++~\cite{PointNet++}-\textit{Pangea}           & 45.79 & 30.52\\ 
        CLIP~\cite{clip}-\textit{Pangea}                       & 48.53 & 32.74\\
        P2S                                           & \textbf{49.69} & \textbf{33.87} \\
    \end{tabular}}
    \vspace{-10px}
    \captionof{table}{Results on the 3D benchmark BABEL-120~\cite{BABEL:CVPR:2021}.} 
    \label{tab:babel}
\end{minipage}
\hfill
\begin{minipage}{0.25\textwidth}
    \centering
    \resizebox{\linewidth}{!}{
    \begin{tabular}{l|c}
        Methods             & Top-1\%  \\
        \toprule[2pt]
        SGN~\cite{tseng2022haa4d}                & 53.3 \\
        PointNet++~\cite{PointNet++}             & 38.6 \\
        CLIP~\cite{clip}                         & 38.0 \\
        \hline
        PointNet++~\cite{PointNet++}-\textit{Pangea}      & 45.6 \\
        CLIP~\cite{clip}-\textit{Pangea}                  & 49.3 \\
        P2S                            & \textbf{54.1}  \\
    \end{tabular}}
    \vspace{-10px}
    \captionof{table}{Results on the 3D benchmark HAA4D~\cite{tseng2022haa4d}.} 
    \label{tab:haa4d}
\end{minipage}
\vspace{-8px}
\end{figure*}

\subsection{Dataset and Implementation}
\label{sec:data_metric_imple}
{\bf Dataset}. 
\textit{Pangea} is adopted to evaluate verb node classification.
We also conduct transfer learning on several multi-modal benchmarks: HICO~\cite{hico}, HAA~\cite{haa500}, 
HMDB51~\cite{hmdb}, Kinetics-400~\cite{kinetics400}, BABEL~\cite{BABEL:CVPR:2021}, and HAA4D~\cite{tseng2022haa4d}. 

{\bf Implementation}.
\textbf{(1)} P2S training: we use 19.5 M 2D images/frames and 840 K 3D humans.
\textbf{(2)} P2S transfer learning: P2S pretrained on \textit{Pangea} with node classification is a knowledgeable backbone. 
To make the transfer learning strict, in pretraining, we \textbf{exclude} the val \& test set data of the downstream dataset from \textit{Pangea} train set. 
Then the pretrained backbone is finetuned and tested on downstream datasets.
For different modalities, we use their corresponding data path.
To make our pipeline efficient, we do not adopt complex temporal encoding and video augmentation. 
Instead, we use \textit{simple} strategies to implement the temporal encoding similar to \cite{actionclip}, \eg, mean pooling, a temporal transformer, average prediction of frames, \etc.
P2S is a multi-modal and lite method that is different from the ad-hoc models for sole-modal tasks. 
Thus we can use it as a \textbf{plug-and-play} method, \ie, fusing it with SOTA models in downstream tasks.
As P2S is trained in much broader semantic coverage on large-scale data, its learned bias is different from ad-hoc models. So P2S is complementary to these SOTA models and can improve their performances in the cooperation.
Moreover, we test different ways to fuse 2D and 3D to mine the potential of multi-modal learning. 
The simplest late fusion (fusing logits) performs best in our tests (Suppl.~Sec.~\textcolor{red}{9}). 
Thus, we use late fusion as the default. 
For data with one human per image/frame, we fuse the 2D and 3D results.
For data with more than one human per image/frame, we first conduct max pooling on the 3D results of multi-human then perform late fusion with 2D.
All experiments are conducted on 4 RTX 3090 GPUs. 

\subsection{Action Recognition}\label{sec:exp_result}

\subsubsection{Verb Node Classification}
To evaluate the verb node classification, we build a \textit{Pangea} test set with 178 K images.
To evaluate few/zero-shot learning, we split the 290 leave nodes into two sets and evaluate them separately: \textit{rare} (133 leave nodes) and \textit{non-rare} (157 leave nodes). 
We report the results in 
Tab.~\ref{tab:node_benchmark}. 
For baseline CLIP, we load the vanilla CLIP pretrained model~\cite{clip} as the backbone and train it on \textit{Pangea} train set for node classification. We use visual-language contrastive learning in training and use the same texts as P2S in inference. 
It achieves 28.25 mAP on 290 leave nodes (16.90 mAP for 133 rare nodes, 37.87 mAP for 157 non-rare nodes). 
Relatively, P2S performs much better with the help of disentanglement and semantic/geometric information. It achieves 34.25 mAP (21.56 for rare nodes, and 45.00 for non-rare nodes). 
Moreover, with label augmentation, P2S-aug further outperforms P2S on all three tracks.

\subsubsection{Transfer Learning} 
We refer to the downstream benchmark as the \textit{target}.
For a fair comparison, we design several baselines: 
(1) CLIP: finetuning the vanilla CLIP pretrained model on the target train set and testing it on the target test set. The output is activity predictions $\mathcal{S}_{act}$, and the loss is contrastive loss $\mathcal{L}_{act}$;
(2) CLIP-\textit{Pangea}: finetuning the vanilla CLIP pretrained model on \textit{Pangea} train set with $\mathcal{L}_{act}$, then finetuning it on the target train set, where $\mathcal{S}_{act}$ is used for evaluation on the target test set.
(3) P2S: detailed in Sec.~\ref{sec:train_infer}, where the output $\mathcal{S}_{act}$ is fused with the better one from CLIP/CLIP-\textit{Pangea}.

{\bf Image Benchmark}.
In Tab.~\ref{tab:hico}, CLIP performs well and even outperforms the ad-hoc SOTA models on HICO~\cite{hico}.
Pretrained on the image-text pairs from \textit{Pangea},
CLIP-\textit{Pangea} is inferior to CLIP because of the large domain gap between activity videos in Pangea and human-object interaction images in HICO~\cite{hico}.
Thus, CLIP-\textit{Pangea} cannot utilize the extensive semantic-geometric knowledge.
Relatively, P2S boosts the performance and outperforms RelViT and CLIP with \textbf{7.62} and \textbf{1.39} mAP respectively.

\begin{table}[t]
    \centering
    \resizebox{0.95\linewidth}{!}{
    \begin{tabular}{c | c | c | c}
        P2S & HICO\cite{hico} mAP & HAA~\cite{haa500} Acc (\%) &  HMDB51~\cite{hmdb} Acc (\%) \\ 
        \toprule[2pt]
        \XSolidBrush  & 41.32 & 68.87 &  70.80 \\
        \CheckmarkBold & \textbf{46.91} & \textbf{70.87} &  \textbf{71.09}
      \end{tabular}}
      \vspace{-9px}
        \caption{Results of P2S $+$ MLLM~\cite{gao2023llama} on several datasets.}
        \label{tab:with_vllm}
        \vspace{-10px}
\end{table}

\begin{figure}[t]
    \centering
    \includegraphics[width=0.95\linewidth]{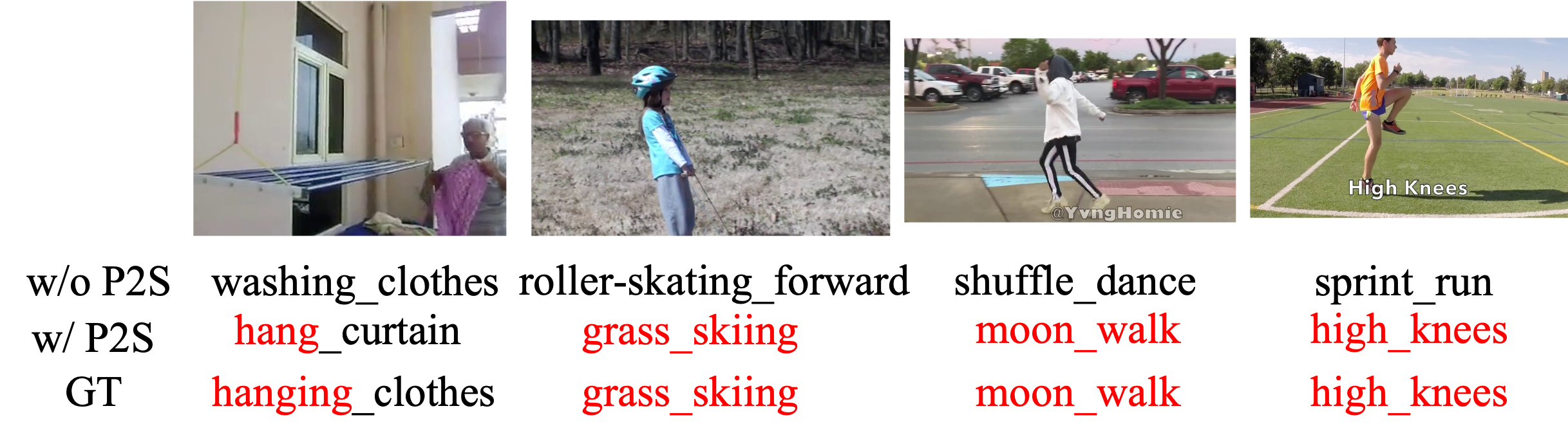}
    \vspace{-7px}
    \caption{Result analysis of MLLM~\cite{gao2023llama} w/ or w/o P2S. 
    } 
    \label{fig:failure_case}
    \vspace{-10px}
\end{figure}

{\bf Video Benchmark}.
The CLIP, CLIP-\textit{Pangea} are with the same setting as above. The conclusion is similar in Tab.~\ref{tab:haa}-\ref{tab:kinetics}. On HAA and HMDB51, CLIP-\textit{Pangea} weaponized with \textit{Pangea} outperforms CLIP. P2S outperforms CLIP with \textbf{5.07}\%, \textbf{0.90}\% and \textbf{0.98}\% respectively on 3 benchmarks respectively. Moreover, P2S without bells and whistles performs comparably well (\eg, TSN on HMDB51, TSN on Kinetics-400) or even better (\eg, TSN on HAA) compared with ad-hoc SOTA. Lastly, fusing P2S and SOTA models further improves the performance: \textbf{4.47}\% (HAA), \textbf{1.41}\% (HMDB51), \textbf{2.58}\% (Kinetics-400). 

\begin{figure}[t]
    \centering
    \includegraphics[width=0.85\linewidth]{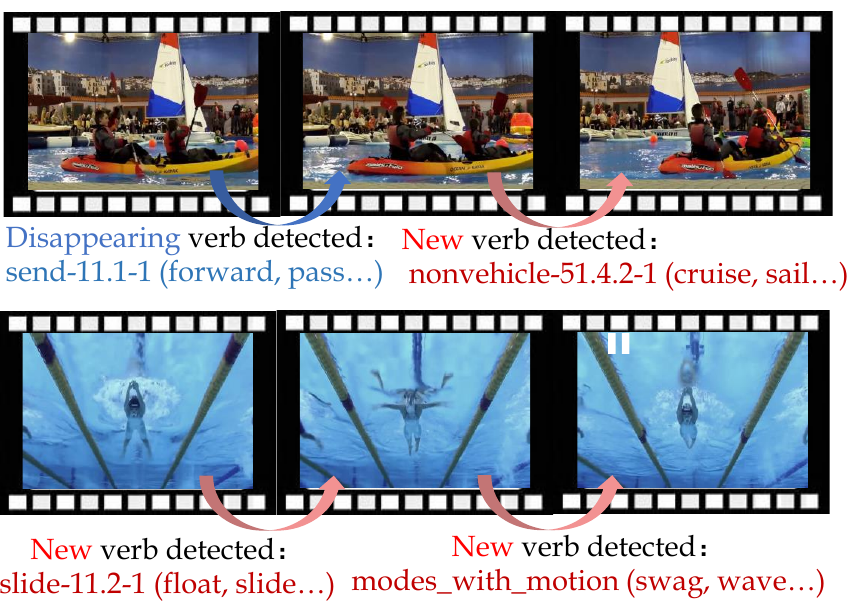} 
    \vspace{-7px}
    \caption{Visualization of changed node predictions from 2 videos. 
    } 
    \label{fig:demo_video}
    \vspace{-10px}
\end{figure}

\begin{figure}[t]
    \centering
    \includegraphics[width=0.95\linewidth]{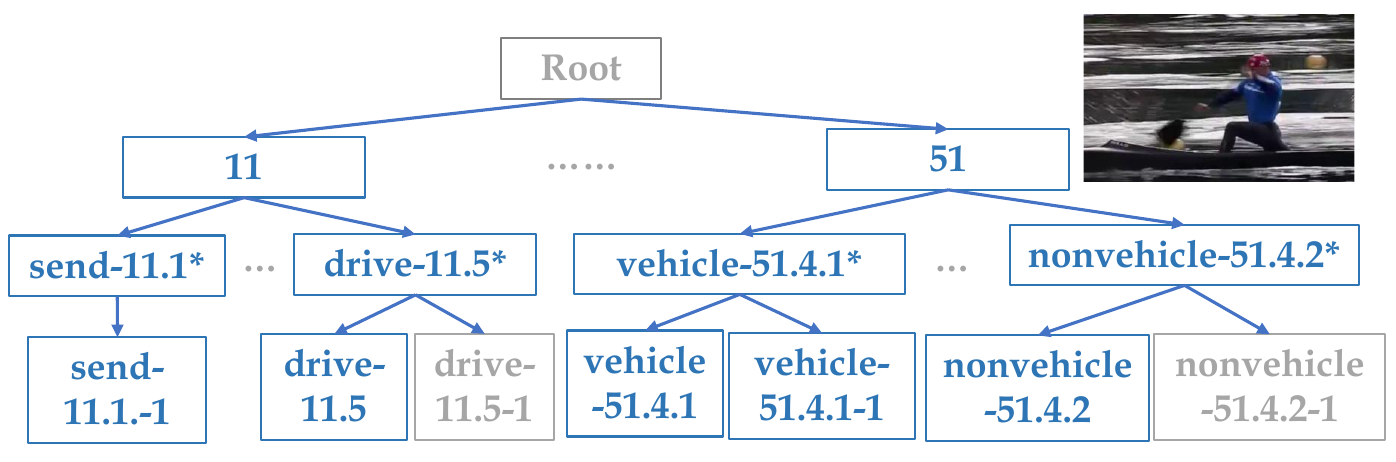} 
    \vspace{-7px}
    \caption{Hierarchical predictions of an image with action \texttt{canoeing\_sprint}. P2S outputs 898 node predictions for the image, and some nodes among the top 15 highest predictions are shown in the blue blocks.
     P2S can learn from generic concepts (\eg, \texttt{11:sending and carrying}) to finer-grained (\eg, \texttt{drive-11.5}) concepts.
    } 
    \label{fig:hierarchical_pred}
    \vspace{-8px}
\end{figure}

{\bf 3D Benchmark.}
We set baselines PointNet++ and CLIP and strengthen them with \textit{Pangea} as PointNet++-\textit{Pangea} and CLIP-\textit{Pangea} (Suppl.~Sec.~\textcolor{red}{7}). 
Similarly, PointNet++-\textit{Pangea} and CLIP-\textit{Pangea} performs better in Tab.~\ref{tab:babel},~\ref{tab:haa4d}.
P2S outperforms all the baselines, \eg \textbf{7.43}\%  upon PointNet++ on BABEL.
Moreover, P2S performs better than the ad-hoc SOTA thanks to the abundant data of \textit{Pangea}.
We do not fuse P2S with SOTA due to the modality gap: most SOTA use 3D skeleton while we use point cloud. 

{\bf Integration with MLLM}.
P2S can facilitate recent powerful Multi-Modal Large Language Models (MLLM) as a plug-and-play method. We integrate the prediction of P2S with a SOTA MLLM backbone: 
LLaMA-Adapter V2~\cite{gao2023llama} on HAA~\cite{haa500}, HICO~\cite{hico} and HMDB~\cite{hmdb}.
When trained without P2S, the backbone is finetuned on the train set to output captions indicating the activity. Then the top-1 accuracy/mAP is calculated by comparing the semantic distance between output captions and GT actions based on a CLIP text encoder. When trained with P2S, the P2S prediction is converted into a prompt as known information for the MLLM. 
The results are shown in Tab.~\ref{tab:with_vllm}. The performance improvement shows the complementary effectiveness of P2S to enhance MLLM.
We also show some cases predicted by MLLM with and without P2S in Fig.~\ref{fig:failure_case}.
In the first column, though MLLM with P2S does not predict the correct action, it does predict the \textbf{correct verb} thanks to the knowledge from \textit{Pangea}, making the prediction semantically similar to the ground truth. In other columns, with the help of P2S, MLLM gives the correct prediction.

\subsection{Further Analysis}
\begin{figure}[t]
    \centering
    \vspace{-7px}
    \includegraphics[width=\linewidth]{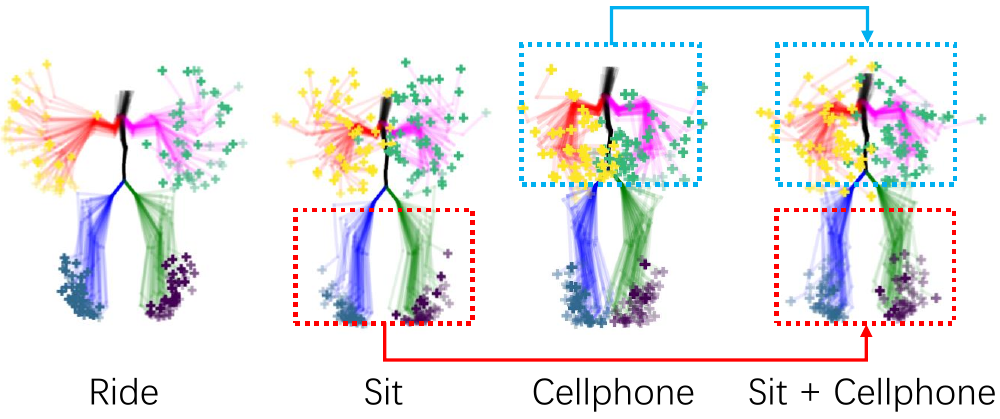}
    \vspace{-15px}
    \caption{S2P results.
    \texttt{ride} has the elbows away from the spine,
    while \texttt{sit} has the opposite.
    Adding \texttt{cellphone} upon \texttt{sit} drives the wrist to distribute around the pelvis more.
    }
    \label{fig:samp_vis}
    \vspace{-8px}
\end{figure}

{\bf Visualization}.
We analyze changed node predictions in videos in Fig.~\ref{fig:demo_video} and show hierarchical predictions in Fig.~\ref{fig:hierarchical_pred}.
P2S effectively captures the subtle semantic changes hierarchically.
Besides, we can also conduct motion generation given \textit{Pangea}, \ie, \textbf{Semantic-to-Physical Space} (\textbf{S2P}), to fully represent its efficacy. 
In Fig.~\ref{fig:samp_vis}, we show the results of inputting verb nodes and use a simple cVAE to generate 3D motions, verifying that S2P is capable of generating reasonable poses for single/multi-node. 

\begin{table}[t]
    \centering
    \resizebox{0.8\linewidth}{!}{
    \begin{tabular}{l | c c c }
        Method  & Full & Rare & Non-Rare \\
        \toprule[2pt]
        P2S-aug & \textbf{34.25}   & \textbf{21.56} & \textbf{45.00}   \\ 
        \hline
        w/o Disentanglement   &  30.09   & 18.65 & 39.79\\ 
        w/o Semantic Augmentation   & 34.01  & 21.37 & 44.72 \\ 
        w/o Text Encoder & 31.81    & 20.05 & 41.78\\ 
        w/o Hyperbolic Mapping & 32.56   & 20.49 & 42.78\\ 
        \hline
        w/o Semantic Prompt & 33.20    & 21.00 & 43.54\\ 
        w/o Geometric Prompt &  33.81    & 21.20 & 44.49 \\ 
      \end{tabular}}
      \vspace{-10px}
      \captionof{table}{Ablation studies on the proposed benchmark \textit{Pangea}.} 
      \vspace{-8px}
    \label{tab:ablation} 
\end{table}

{\bf Performance Variance Analysis.} 
a) 3D vs. 2D: 
P2S presents more evident performance improvement on 3D benchmarks because of the smaller-scale train/test set and smaller data domain gaps than 2D image/video benchmarks.
b) Image vs. Video Benchmark:
P2S performs better in image benchmark as the baseline is a concise, image-based model, rather than a sophisticated video-based one.
c) Variations within Video Benchmarks:
P2S shows different benefits across video benchmarks mainly because of various sizes of pre-training data and node sample distribution.

{\bf Ablation Study \& Discussion}.
We conduct ablations on \textit{Pangea} to evaluate the P2S components in Tab.~\ref{tab:ablation}.
Without four key components, P2S shows obvious degradation, which follows the gap between P2S and CLIP-\textit{Pangea}. 
Moreover, semantic disentanglement matters most to facilitate the one-to-many P2S mapping and weakly-supervised learning.
Here, we adopt concise models to verify the efficacy of \textit{Pangea} and quickly trial-an-error with limited GPUs. We believe that larger and more sophisticated models trained with \textit{Pangea} with more computing power would gain more superiority in future work.
For additional results and discussions, please also refer to the supplementary.

\section{Conclusion}
In this work, to bridge the action data ``isolated islands'', 
we propose a structured semantic space and accordingly merge multi-modal datasets into a unified \textit{Pangea}. 
Moreover, to fully use \textit{Pangea}, we propose a concise mapping system to afford multi-modal action recognition showing superiority.
We believe our framework paves a new path for future study.

\section{Acknowledgments} 
This work is supported in part by the
National Natural Science Foundation of China under Grants 62306175, 
National Key Research and Development Project of China (No. 2022ZD0160102, No. 2021ZD0110704), 
Shanghai AI Laboratory, XPLORER PRIZE grants.

{
    \small
    \bibliographystyle{ieeenat_fullname}
    \bibliography{main}
}

\clearpage

We report more details and discussions here:

Sec.~\ref{sec:more_result}: Supplementary Related Works

Sec.~\ref{sec:detail_pangea}: Details of \textit{Pangea} Database

Sec.~\ref{sec:detail_p2s}: Details of P2S

Sec.~\ref{sec:detail_s2p}: Details of S2P

Sec.~\ref{sec:detail_exper}: Datasets Details in Experiments

Sec.~\ref{sec:detail_analysis_transfer}: Details of Image/Video Transfer Learning

Sec.~\ref{sec:detail_analysis_transfer_3d}: Details of 3D Transfer Learning

Sec.~\ref{sec:add_result}: Additional Results of P2S and S2P

Sec.~\ref{sec:add_ablation}: Additional Ablation Studies

Sec.~\ref{sec:discussion}: More Discussions

\section{Supplementary Related Works}
\label{sec:more_result}
\subsection{Hyperbolic Representation}
Hyperbolic representation has emerged in deep learning to encode hierarchical tree-like structure and taxonomy~\cite{nickel2017poincare,tifrea2018poincar,ganea2018hyperbolic}.
It has been applied in computer vision for hierarchical action search~\cite{hyperbole}, video action prediction~\cite{suris2021learning}, and hierarchical image classification~\cite{khrulkov2020hyperbolic,dhall2020hierarchical,liu2020hyperbolic}. 
Long~\etal~\cite{hyperbole} project video and action embeddings in the hyperbolic space and train a cross-modal model to perform hierarchical action search. 
In this work, we use hyperbolic embeddings to encode the hierarchical geometry of our structured semantic space.

\subsection{Visual-Language Learning} 
Visual-language learning recently shows potential in learning generic representations~\cite{clip,jia2021scaling,singh2022flava,yuan2021florence,wang2021simvlm}.
Specifically, CLIP~\cite{clip} and ALIGN~\cite{jia2021scaling} benefit from web-scale curated image-text pairs for training and allow zero-shot transfer to many downstream tasks.
Following works~\cite{actionclip,lin2022frozen} adapt CLIP to video recognition via prompting, temporal modeling, \etc.
However, it may be hard for their implicit language embedding to capture the subtle taxonomy and structure knowledge of action semantics. 
Thus, we propose to solve the problem via a structured semantic space.

\subsection{3D Human Representation}
3D Human Representation has been attracting much attention for a long time.
A most intuitive representation is the 3D human pose, and lots of effort has been put into single-view 3D pose reconstruction~\cite{pavlakos2017coarse,li2020hmor,martinez2017simple,sun2017compositional}. 
Some methods~\cite{pavlakos2017coarse,sun2017compositional} directly regress 3D pose from the given image. 
With great progress given in 2D pose estimation, many works~\cite{li2020hmor,martinez2017simple} adopt pre-detected 2D poses as auxiliary inputs.
DensePose~\cite{densepose} proposes to adopt a UV map to represent the dense correspondence between the image and a human mesh, which could function as a 2.5D human representation.
Lately, different parametric human body models (like SMPL~\cite{loper2015smpl} and SMPL-X~\cite{smplify-x}) are proposed as promising human representations. Impressive performance has been achieved with weak supervision, like 2D pose~\cite{kanazawa2018end,loper2015smpl,smplify-x,expose,romp}, semantic segmentation, motion dynamics~\cite{kp}, and so on. Also, different paradigms are proposed. Some works~\cite{SMPL,SMPL-X} directly fit the parametric model to the weak supervision signals, which is accurate but sensitive to the initial state, and the speed is restricted.
While there are also regression methods~\cite{kanazawa2018end,expose,romp} learning a neural network to map images to human model parameters, greatly accelerating the reconstruction but losing accuracy.
Combining the advantages of both kinds of methods, SPIN~\cite{spin} and EFT~\cite{eft} proposed to adopt regression methods for initialization and then use fitting methods for refinement.
Inspired by the recent progress in NeRF~\cite{mildenhall2020nerf}, HumanNeRF~\cite{weng_humannerf_2022_cvpr} proposed a neural radiance field representation for free-view dynamic human modeling. 

\subsection{3D Action Generation} 
3D Action Generation is an active field.
With large skeleton datasets such as NTU~\cite{liu2019ntu} and Human3.6M~\cite{human3}, considerable efforts have been put on it~\cite{action2motion,petrovich21actor,petrovich22temos,tevet2022motionclip,zhang2022motiondiffuse}.
Meanwhile, MoCap datasets~\cite{AMASS:ICCV:2019,action2motion} push it further towards parametric human model-based generation~\cite{petrovich21actor}.
Most efforts are either unconditional or conditioned on restricted action classes.
Beyond class conditioned generation, some works conduct generation with natural language~\cite{petrovich22temos,tevet2022motionclip} based on datasets composed of motion-text pairs~\cite{BABEL:CVPR:2021,kit}.

\section{Details of \textit{Pangea} Database}
\label{sec:detail_pangea}
\subsection{Data Curation}
With the structured semantic space, 
We can collect data with diverse modalities, formats, and granularities, and adapt them into a unified form. 
Our database \textit{Pangea} contains a large range of data including images, videos, and skeletons/MoCaps. 
We give more details of the processing and formulation as follows:
    
\textbf{1) Semantic consistency.} 
The class definitions of datasets are various, but they can be mapped to our semantic space with the fewest semantic damages. 
As mentioned in the main text, the mapping is completed via manual annotation with the help of word embedding~\cite{clip} distances and OpenAI GPT-3.5. Manual annotation is the most accurate and most expensive, while word embedding comparison is the least. Thus, we adopt a hybrid method: potential class-node mapping is first filtered out roughly by comparing word embedding, then selected via GPT-3.5 prompting, and finally checked by human annotators. As more and more classes are aligned and covered, the process would be faster and faster with synonyms checking. 

\begin{figure}[t]
    \centering
    \includegraphics[width=\linewidth]{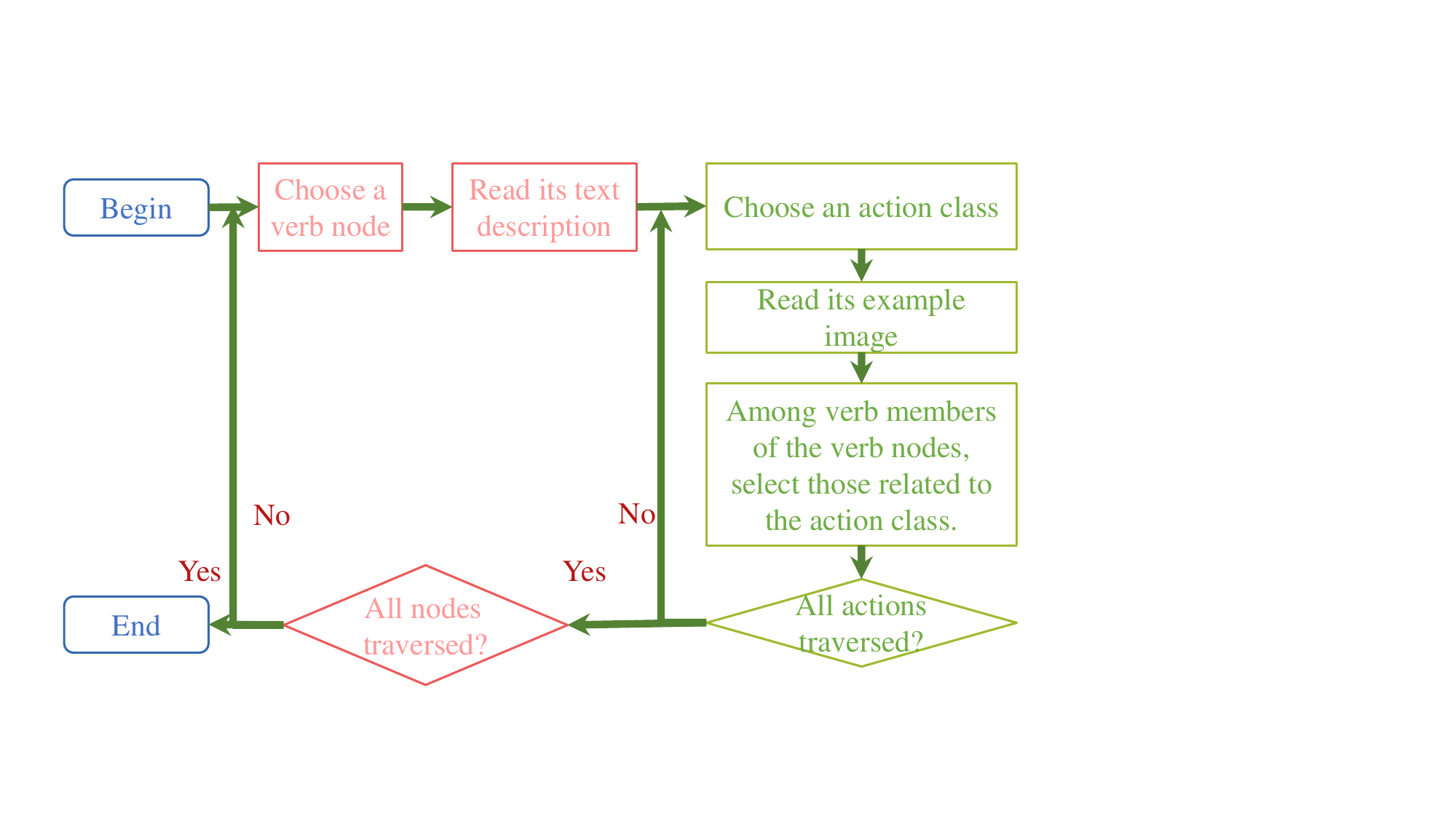}
    \caption{The flow chart of the action semantic mapping by human annotators.}
    \label{fig:label_alignment}
\end{figure}

Suppl.~Fig.~\ref{fig:label_alignment} shows a flow chart of the action semantic mapping by human annotators.
We invite 60 annotators of different backgrounds. Each candidate class is annotated three times, generating the final labels via the majority rule. Finally, for the 898 verb nodes (including 575 leaf nodes), there are a total of 515 verb nodes that have corresponding retargeted classes (including 290 leaf nodes). The missing verb nodes are mostly related to visually unrecognizable semantics, \eg, \texttt{invest}.

\textbf{2) Temporal consistency.} 
Some videos~\cite{kinetics-700} only have \textit{sparse} labels for a whole clip instead of each frame. 
To solve this conflict, we sample the clip with \textit{3 FPS} and give them the label of their belonged clip describing the action in the clip.
More dense or spare sampling is either computationally costly or with serious information loss. 
On the contrary, with \textit{dense} frame labels~\cite{AVA}, we can easily get the clip label via fusing frame labels.
Thus, we provide both frame- and clip-level labels for videos.
    
\textbf{3) Spatial consistency.} 
There are both instance (boxes)~\cite{hicodet} and image~\cite{hico} level labels. 
It is too expensive to annotate all missing human boxes and actions to make the whole \textit{Pangea} instance-level. 
More realistically, we merge the instance labels of each image/frame into image/frame labels.
In the future, we can also add more box labels to existing images based on the existing instance labels to support larger-scale instance-level training.
    
\textbf{4) 3D format consistency.} 
3D action datasets typically have different formats, \eg, SMPL~\cite{SMPL} contains $24$ key-points while CMU MoCap~\cite{action2motion} has $31$ key-points. To keep format consistency, we transform all of them into SMPL via a fitting procedure. 
    
\textbf{5) 2D-3D consistency.} Image/video datasets mostly contains only 2D labels without 3D human labels. We generate 3D humans via single-view reconstruction~\cite{romp}. 
Please refer to Suppl.~Sec.~\ref{sec:3drecon} for more details.

\subsection{3D Human Body Annotation Details}
\label{sec:3drecon}
We adopt 3D humans for multiple reasons.
First, 3D human provides a robust representation without \textit{viewpoint} problems.
Second, 3D humans can be seen as the safest choice as the physical carrier of actions with no need to consider the \textit{domain gap} across image conditions.

In \textit{Pangea}, we also prepare pseudo 3D human labels for images/videos.
Different strategies are adopted depending on the label circumstances of the data. 
For different scenarios with ground truth (GT) 2D or 3D human poses, human boxes only, and no human instance labels at all, we adopt different strategies as follows:

\begin{figure*}[t]
    \centering
    \includegraphics[width=0.9\linewidth]{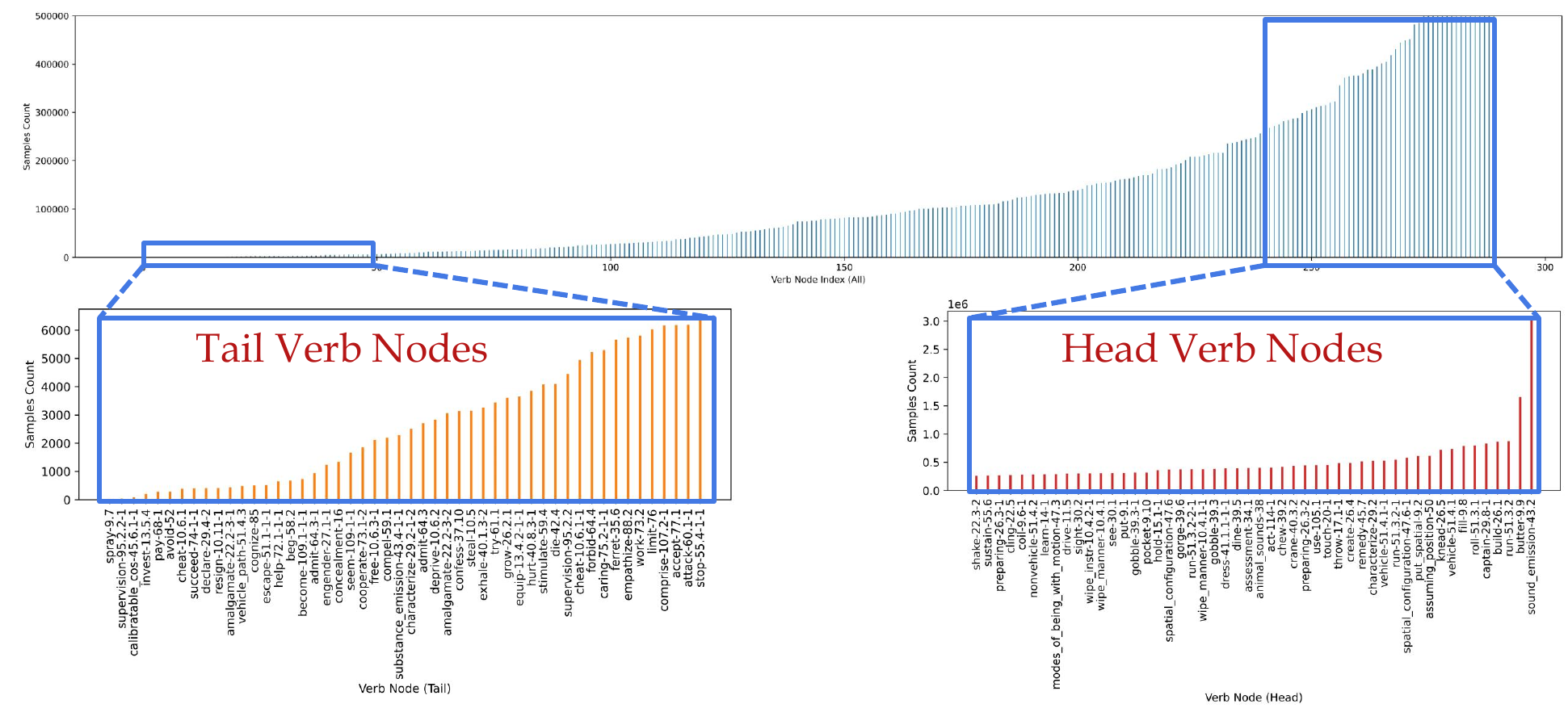}
    \caption{Semantic distribution of samples on 290 leave nodes, including detailed statistics on tail/head verb nodes.}
    \label{fig:semantic_distribution}
\end{figure*}

\begin{enumerate}
    \item If an image has a 3D human pose annotation, we fit the SMPL model to the 3D pose and associate the fitted 3D human body with the annotation. The 2D body is acquired by cropping the image with the bounding box.
    \item If an image has a 2D human pose annotation, we calculate the MSE error of the annotated pose and the re-projected pose from 3D recovering and associate the annotated human instance with the reconstructed 3D body whose MSE error is the lowest among all and lower than a threshold. The 2D body representation is acquired by cropping the image with the box.
    \item If an image has a human bounding box annotation, we calculate the IoU between the annotated box and the re-projected human mesh bounding box. Then, the annotated human box is associated with the 3D human body whose IoU is the highest and higher than a threshold. The 2D body representation is acquired by cropping the image with the bounding box.
    \item If an image contains no human annotation, OpenPose~\cite{openpose} is adopted to generate a pseudo annotation for the 2D human pose. Then we follow the same association strategy as images with 2D pose annotation. We assume the human instance with the lowest MSE error is the target human performing the annotated action.
    \item For mesh sequences, we directly adopt them as 3D humans. Besides, for skeleton sequences without a 2D image available, we align the annotations with joints defined by SMPL and extract the 3D human body by fitting the SMPL model to the aligned pose.
\end{enumerate}

Note that the 3D human pose and the corresponding/re-projected 2D pose could be easily extracted simultaneously. 
Images/frames with no human bodies or failure reconstructions were dropped.
In practice, ROMP~\cite{romp} and EFT~\cite{joo} are adopted to directly recover humans from images.

\subsection{More Statistics of \textit{Pangea}}
We list the collected datasets of \textit{Pangea} in Suppl.~Tab.~\ref{tab:dataset_statistics}.
\begin{table*}[t]
    \centering
    \resizebox{0.65\linewidth}{!}{
    \begin{tabular}{ll|c|c|c}
    \hline
    &       & Action Classes & Images/Frames &  Videos \\ 
    \hline
    \multicolumn{1}{l|}{\multirow{7}{*}{Image}} & Willow Action~\cite{willow}   &  7   & 1 K    &  -  \\ 
    \cline{2-5} \multicolumn{1}{l|}{} & Phrasal Recognition~\cite{VisualPhrases} &  10  & 4 K    &  -  \\ 
    \cline{2-5} \multicolumn{1}{l|}{} & Stanford 40 Actions~\cite{stanford40}    &  40  & 4 K    &  -  \\ 
    \cline{2-5} \multicolumn{1}{l|}{} & MPII~\cite{mpii}                         &  410 & 4 K   &  -  \\ 
    \cline{2-5} \multicolumn{1}{l|}{} & HICO~\cite{hico}                  &  600 & 38 K   &  -  \\ 
    \cline{2-5} \multicolumn{1}{l|}{} & HAKE~\cite{pastanet}                     &  156 & 42 K  &  -  \\ 
    \hline
    \multicolumn{1}{l|}{\multirow{14}{*}{Video}} & HMDB51~\cite{hmdb}           &  51  & 69 K   &  7 K    \\ 
    \cline{2-5} \multicolumn{1}{l|}{} & HAA500~\cite{haa500}                     &  500 & 64 K  &  8.5 K       \\ 
    \cline{2-5} \multicolumn{1}{l|}{} & AVA~\cite{AVA}                           &  80  & 162 K  &  0.5 K     \\ 
    \cline{2-5} \multicolumn{1}{l|}{} & YouTube Action~\cite{youtube_action}     &  11  & 4 K    &  1 K      \\ 
    \cline{2-5} \multicolumn{1}{l|}{} & ASLAN~\cite{ASLAN}                       &  432 & 18 K    &  1 K       \\ 
    \cline{2-5} \multicolumn{1}{l|}{} & UCF101~\cite{ucf101}                     &  101 & 61 K  &  13 K     \\ 
    \cline{2-5} \multicolumn{1}{l|}{} & Olympic Sports~\cite{olympic_sports}     &  16  & 6 K   &  1 K      \\ 
    \cline{2-5} \multicolumn{1}{l|}{} & Penn Action~\cite{penn_action}           &  15  & 85 K   &  2 K      \\ 
    \cline{2-5} \multicolumn{1}{l|}{} & Charades~\cite{charades}                 &  157 & 44 K   &  8 K     \\ 
    \cline{2-5} \multicolumn{1}{l|}{} & Charades-Ego~\cite{charades_ego}         &  157 & 235 K  &  8 K     \\ 
    \cline{2-5} \multicolumn{1}{l|}{} & ActivityNet~\cite{caba2015activitynet}   &  200 & 2,444 K &  20 K     \\ 
    \cline{2-5} \multicolumn{1}{l|}{} & HACS~\cite{zhao2019hacs}                 &  200 & 1,379 K &  504 K       \\ 
    \cline{2-5} \multicolumn{1}{l|}{} & Home Action Genome~\cite{home_action_genome} & 453 & 702 K & 6 K      \\ 
    \cline{2-5} \multicolumn{1}{l|}{} & Kinetics~\cite{kinetics-700}             &  700 & 14,132 K &  536 K      \\ 
    \hline
    \multicolumn{1}{l|}{Image+Video} & Pangea & \textbf{4,296}   &    \textbf{19,495 K}    &   \textbf{1,116 K} \\
    \hline
    \multicolumn{1}{l|}{\multirow{7}{*}{Skeleton/MoCap}} & HumanAct12~\cite{action2motion} & 12  & 90 K   & 1 K   \\ 
    \cline{2-5} \multicolumn{1}{l|}{}             & CMU MoCap~\cite{action2motion}  & 8   & 978 K  & 1 K   \\ 
    \cline{2-5} \multicolumn{1}{l|}{}             & UTD-MHAD~\cite{utd}             & 27  & 90 K   & 1 K   \\ 
    \cline{2-5} \multicolumn{1}{l|}{}             & NTU RGB+D~\cite{ntu}            & 120 & 830 K & 114 K   \\ 
    \cline{2-5} \multicolumn{1}{l|}{}             & Human3.6M~\cite{human3}         & 17  & 3,600 K & $<$1 K  \\ 
    \cline{2-5} \multicolumn{1}{l|}{}             & BABEL~\cite{BABEL:CVPR:2021}    & 260 & 4,050K & 10K  \\ 
    \cline{2-5} \multicolumn{1}{l|}{}             & HAA4D~\cite{tseng2022haa4d}     & 300 & 212K & 3K  \\ 
    \hline
    \multicolumn{1}{l|}{Total} & Pangea & \textbf{5,040}   &    \textbf{29,345 K}    &   \textbf{1,247 K} \\
    \hline
    \end{tabular}}
    \caption{Statistics of collected and curated multi-modal datasets. Note that different datasets may share part of 
    action classes (\eg, ActivityNet~\cite{caba2015activitynet} and HACS~\cite{zhao2019hacs}).
    }
    \label{tab:dataset_statistics}
\end{table*}

\subsection{Justifications of Database Design Choices}
The class-node mapping is selected via \textbf{GPT-3.5 prompting}. We choose GPT-3.5 because of its excellent instruction-following abilities and easy-to-use API. In future work, we will extend our work with more powerful LLMs (\eg, GPT-4) or locally deployed LLMs (\eg, Llama 2). 

\textbf{SMPL}. We choose SMPL mainly for its versatility and expressiveness since most data either provide SMPL parameters or could be conveniently converted into SMPL format.

\textbf{ROMP \& EFT}. We find ROMP with EFT optimization managing to process our massive data efficiently with promising reconstruction quality. We would keep refining the data quality with the progress in 3D human reconstruction on our open-sourced website.

\subsection{Semantic Distribution of \textit{Pangea}}
Suppl.~Fig.~\ref{fig:semantic_distribution} shows the sample count for 290 leaf verb nodes of our \textit{Pangea} database. Detailed statistics on tail/head verb nodes are also listed.

\subsection{Data License/Address} 
All the data of \textit{Pangea} are from open-sourced datasets and for research purposes only. We give the data licenses and links of the gathered datasets here.
\begin{itemize}
    \item Willow Action: \url{https://www.di.ens.fr/willow/research/stillactions/}
    \item Phrasal Recognition: \url{https://vision.cs.uiuc.edu/phrasal/}
    \item Stanford 40 Action: \url{http://vision.stanford.edu/Datasets/40actions.html}
    \item MPII: \url{http://human-pose.mpi-inf.mpg.de/}
    \item HICO: \url{http://www-personal.umich.edu/~ywchao/hico/}
    \item V-COCO: \url{https://www.v7labs.com/open-datasets/v-coco}
    \item HAKE: \url{http://hake-mvig.cn/download/}
    \item HMDB51: \url{https://creativecommons.org/licenses/by/4.0}
    \item HAA500: \url{https://www.cse.ust.hk/haa/LICENSE}
    \item AVA: \url{https://creativecommons.org/licenses/by/4.0}
    \item Youtube Action: \url{http://www.cs.ucf.edu/~liujg/YouTube_Action_dataset.html}
    \item ASLAN: \url{https://talhassner.github.io/home/projects/ASLAN/ASLAN-main.html}
    \item UCF101: \url{https://www.crcv.ucf.edu/data/UCF101.php}
    \item Olympic Sports: \url{http://vision.stanford.edu/Datasets/OlympicSports/}
    \item Penn Action: \url{http://dreamdragon.github.io/PennAction/}
    \item Charades: \url{http://vuchallenge.org/license-charades.txt}
     \item Charades-Ego: \url{https://prior.allenai.org/projects/data/charades-ego/license.txt}
     \item ActivityNet: \url{http://activity-net.org/download.html}
     \item HACS: \url{http://hacs.csail.mit.edu/}
     \item Home Action Genome: \url{https://homeactiongenome.org/index.html#what-we-do}
    \item Kinetics: \url{https://creativecommons.org/licenses/by/4.0}
    \item HumanAct12: \url{https://github.com/EricGuo5513/action-to-motion}
    \item CMU MoCap: \url{http://mocap.cs.cmu.edu/}
    \item UTD-MHAD: \url{https://personal.utdallas.edu/~kehtar/UTD-MHAD.html}
    \item NTU RGB+D: \url{https://rose1.ntu.edu.sg/dataset/actionRecognition/}
    \item Human3.6M: \url{http://vision.imar.ro/human3.6m/eula.php}
    \item BABEL: \url{https://babel.is.tue.mpg.de/license.html}
    \item HAA4D: \url{https://cse.hkust.edu.hk/haa4d/}
\end{itemize}

\section{Details of P2S}
\label{sec:detail_p2s}

\subsection{Label Augmentation Details}
We detail the label augmentation here.
Each image has a partial annotation $Y=\{y_i|y_i=1,0,\emptyset\}_{i=1}^N$, where $1,0$ are certain positive/negative labels, and $\emptyset$ are uncertain ones.

\begin{figure}[t]
    \centering
    \includegraphics[width=\linewidth]{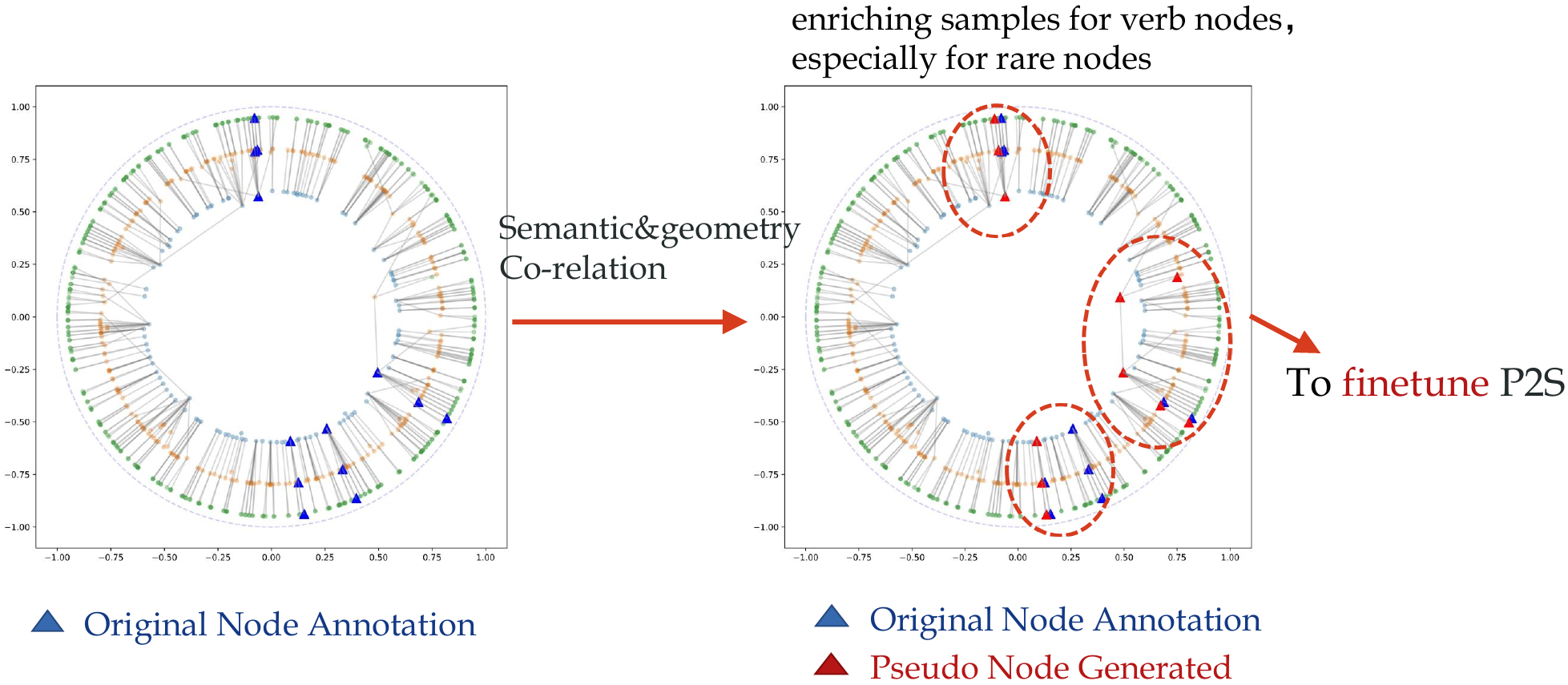}
    \caption{Illustration of label augmentation. Pseudo labels are generated based on VerbNet semantic/geometry co-relation. With generated pseudo labels, we can finetune P2S with more samples, which especially benefits verb nodes with rare samples.}
    \label{fig:pseudo}
\end{figure}

A direct way to solve the uncertain labels is \textit{assuming negative}: unobserved labels are considered as negatives. That is, for $\forall i$, if $y_i=\emptyset$, assign $y_i=0$.
However, some positive labels are falsely treated negatively, which hinders semantic learning, especially for few-shot nodes. 
Therefore, we propose to generate pseudo labels for uncertain labels, instead of simply treating them as negatives. 
That is, if $y_i=\emptyset$, assign $y_i=y_i^{'}\in[0,1]$.
The pseudo label $y_i^{'}$ is generated based on the structure and language prior to our semantic space.
The pre-defined geometry and semantic information in VerbNet indicate the co-relation between verb nodes. Based on the co-relation, high-quality nodes with more samples can transfer knowledge (positive/negative labels) to low-quality nodes with fewer samples, thus generating pseudo labels to apply label augmentation and facilitate P2S learning.
The process is illustrated in Suppl.~Fig.~\ref{fig:pseudo}.

In the implementation, we first obtain a co-relation matrix $\mathbf{C}=\{c_{ij}\}_{N\times N}$ of N verb nodes via language priors and VerbNet structure. Then pseudo labels are generated based on $\mathbf{C}$ and certain labels. That is, for each $i$ where $y_i=\emptyset$, we assign $y_i^{'}=\sum\limits_{j:y_j=1,j \neq i} c_{ij} y_j $.

The co-relation matrix $\mathbf{C}$ is calculated from two components: 
1) $C_{L}$ based on language priors;
2) $C_{E}$ based on VerbNet structure.
For $C_{L}$, we encode the semantic information of each verb node into $l_i$ via a pretrained text encoder~\cite{gao2021simcse} and then construct $C_{L}={cos(l_i, l_j)}$, where $cos(\cdot, \cdot)$ measures the cosine similarity of two vectors.
For $C_{E}$, based on the trained hyperbolic embeddings $E=\{e_i\}_{i=1}^N$, we obtain $C_{E}= -d_{\mathcal{L}}(e_i, e_j)$, where $d_{\mathcal{L}}(\cdot, \cdot)$ is the Lorentzian Distance (detailed in Suppl.~Sec.~\ref{sec:lorentz}).
Finally, we normalize both $C_{L}$ and $C_{E}$ into $[0,1]$ and obtain $C$ via $C=(C_{L}+C_{E})/2$.

\begin{figure*}[t]
    \centering
    \includegraphics[width=\linewidth]{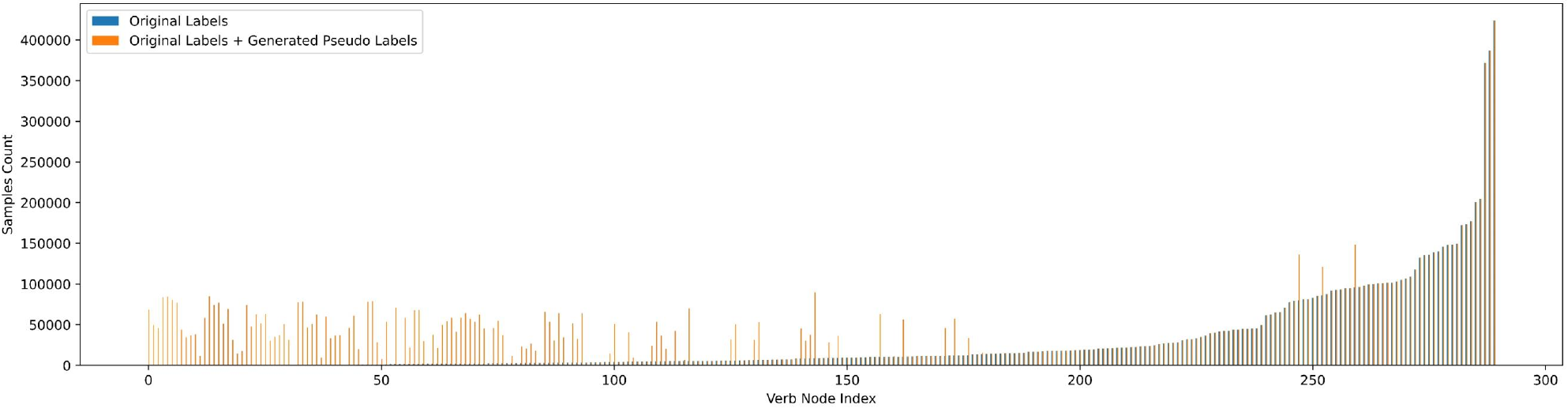}
    \caption{Sample distribution before/after generating pseudo labels.}
    \label{fig:pseudo_samples}
\end{figure*}

With label augmentation, the long-tail distribution is effectively alleviated with credible pseudo labels. The sample distribution before/after generating pseudo labels is shown in Suppl.~Fig.~\ref{fig:pseudo_samples}. We can find that many tail nodes have more samples after the augmentation which alleviates the long-tailed distribution a lot.
To benefit from label augmentation, we train P2S mapping in two phases. 
In phase 1, the whole model is trained via \textit{assuming negative}.
In phase 2, we finetune the model with certain labels and pseudo labels. Phase 2 benefits from the eased long-tail distribution, thus facilitating P2S learning.

Another consideration is to bind prediction with \textbf{soft} or \textbf{hard} pseudo labels. For soft labels, we directly use the pseudo label $y_i^{'}\in [0,1]$ as ground truth. For hard labels, we consider only pseudo labels above the given threshold and use $y_i^{'}\in \{0,1\}$ as ground truth.
We find hard labels drag the performance a little, possibly because of the amplified noise of generated pseudo labels.
Thus, we adopt soft labels in practice.

\subsection{Lorentz Model for Verb Hierarchy}\label{sec:lorentz}

\textbf{Preliminaries~\cite{desai2023hyperbolic}.} 
Lorentz model represents a hyperbolic space of $n$ dimensions on the upper half of a two-sheeted hyperboloid
in $\mathbf{R}^{n+1}$. 
We refer to the hyperboloid’s axis of symmetry as time dimension and all other axes as space dimensions~\cite{desai2023hyperbolic}. Every vector $\mathbf{x} \in  \mathbf{R}^{n+1}$ can be written as $[\mathbf{x}_{\text{space}}, \mathbf{x}_{\text{time}}]$, where $\mathbf{x}_{\text{space}} \in \mathbf{R}^n$ and $\mathbf{x}_{\text{time}} \in \mathbf{R}$.

Let \(\langle \cdot, \cdot \rangle\) be Euclidean inner product and \((\cdot, \cdot)_\mathcal{L}\) denote the \textit{Lorentzian inner product} that is induced by the Riemannian metric of the Lorentz model. For two vectors \(\mathbf{x}, \mathbf{y} \in \mathbb{R}^{n+1}\), it is computed as follows:
\begin{equation}
    (\mathbf{x}, \mathbf{y})_\mathcal{L} = \langle \mathbf{x}_{\text{space}}, \mathbf{y}_{\text{space}} \rangle - \mathbf{x}_{\text{time}} \mathbf{y}_{\text{time}}.
\end{equation}
The induced \textit{Lorentzian norm} is \(\|x\|_\mathcal{L} = \sqrt{(\mathbf{x}, \mathbf{x})_\mathcal{L}}\). The Lorentz model possessing a constant curvature \(-c\) is defined as the following set of vectors:
\begin{equation}
    \mathcal{L}^n = \{\mathbf{x} \in \mathbb{R}^{n+1}: (\mathbf{x}, \mathbf{x})_\mathcal{L} = -1/c, \ c > 0\}.
\end{equation}
All vectors in this set satisfy the following constraint:
\begin{equation}
    \mathbf{x}_{\text{time}} = \sqrt{1/c + \|\mathbf{x}_{\text{space}}\|^2}.
    \label{eq:x_space_time}
\end{equation}

\textbf{Lifting Embeddings onto the Hyperboloid~\cite{desai2023hyperbolic}.} 
We map the physical representation $V=\{v_i\}_{i=1}^N$ and node representation $E=\{e_i\}_{i=1}^N$ into the Lorentz model as $v_i^{\mathcal{L}}$ and $e_i^{\mathcal{L}}$ via the exponential map.
Let the embedding vector ($v_i, g_i$) be \(\mathbf{v}_{\text{enc}} \in \mathbb{R}^n\). We need to apply a transformation such that the resulting vector lies on the Lorentz hyperboloid \(\mathcal{L}^n\) in \(\mathbb{R}^{n+1}\). Let the vector \(\mathbf{v} = [\mathbf{v}_{\text{enc}}, 0] \in \mathbb{R}^{n+1}\). 
We parameterize \textit{only the space components} of the Lorentz model (\(\mathbf{v}_{\text{enc}} = \mathbf{v}_{\text{space}}\))~\cite{desai2023hyperbolic}. Due to such parameterization, we can simplify the exponential map as:
\begin{equation}
\mathbf{x}_{\text{space}} = \frac{\sinh(\sqrt{c \| \mathbf{v}_{\text{space}} \|})}{\sqrt{c \| \mathbf{v}_{\text{space}} \|}} \mathbf{v}_{\text{space}}.
\end{equation}
The corresponding \textit{time component} \(x_{\text{time}}\) can be computed from \(\mathbf{x}_{\text{space}}\) using Eq.~\ref{eq:x_space_time}. The resulting \(\mathbf{x}\) always lies on the hyperboloid.
To prevent numerical overflow, we scale all vectors $\mathbf{v}_{\text{space}}$ in a batch before applying the mapping using two learnable scalars $\omega_{img}$ and $\omega_{txt}$. These are initialized to $\sqrt{1/n}$ so that the Euclidean embeddings have an expected unit norm at initialization.

\textbf{Lorentzian Distance~\cite{desai2023hyperbolic}.} 
The similarity $\mathcal{S}(v_i, e_i)$ is measured via the negative of Lorentzian distance $d_{\mathcal{L}}(\cdot, \cdot)$ between $v_i^{\mathcal{L}}$ and $e_i^{\mathcal{L}}$. 
A \textit{geodesic} is the shortest path between two points on the manifold. Geodesics in the Lorentz model are curves traced by the intersection of the hyperboloid with hyperplanes passing through the origin of \(\mathbb{R}^{n+1}\). The \textit{Lorentzian distance} between two points \(\mathbf{x}, \mathbf{y} \in \mathcal{L}^n\) is:
\begin{equation}
d_\mathcal{L}(\mathbf{x}, \mathbf{y}) = \sqrt{\frac{1}{c} \cdot \cosh^{-1}(-c (\mathbf{x}, \mathbf{y})_\mathcal{L})}.
\end{equation}

\textbf{Entailment Cone~\cite{desai2023hyperbolic}.}
If $y_i=1$, the physical representation $v_i^{\mathcal{L}}$ should lie inside the entailment cone~\cite{ganea2018hyperbolic} of the node representation $e_i^{\mathcal{L}}$.

For each \(\mathbf{x}\), which narrows as we go farther from the origin, the entailment cone is defined by the half-aperture:
\begin{equation}
\alpha(\mathbf{x}) = \sin^{-1}\left(\frac{2K}{\sqrt{c \|\mathbf{x}_{\text{space}}\|}}\right),
\end{equation}
where a constant \( K = 0.1 \) is used for setting boundary conditions near the origin. 
We aim to identify and penalize occasions where the paired image embedding \(\mathbf{y}\) lies outside the entailment cone. For this, we measure the exterior angle \(\theta(\mathbf{x}, \mathbf{y}) = \pi - \angle O\mathbf{x}\mathbf{y}\):
\begin{equation}
\theta(\mathbf{x}, \mathbf{y}) = \cos^{-1}\left(\frac{y_{\text{time}} + x_{\text{time}} c (\mathbf{x}, \mathbf{y})_{\mathcal{L}}}{\|\mathbf{x}_{\text{space}}\| \sqrt{(c (\mathbf{x}, \mathbf{y})_{\mathcal{L}})^2 - 1}}\right).
\end{equation}

If the exterior angle is smaller than the aperture, then the partial order relation between \(\mathbf{x}\) and \(\mathbf{y}\) is already satisfied and we need not penalize anything, while if the angle is greater, we need to reduce it. This is captured by the following loss function (written below for a single \(\mathbf{x}, \mathbf{y}\) pair):
\begin{equation}
\mathcal{L}_{\text{entail}}(\mathbf{x}, \mathbf{y}) = \max(0, \theta(\mathbf{x}, \mathbf{y}) - \alpha(\mathbf{x})).
\end{equation}

\section{Details of S2P}
\label{sec:detail_s2p}
Though we focus on P2S mapping, with the learned abundant semantic representation of nodes and the collected 3D data, we wonder if we can do the inverse mapping, \ie, Semantic-to-Physical space (S2P).
S2P should be scalable to different semantic granularities and flexible with either \textbf{single}- or \textbf{multi}-node and generate reasonable 3D motions.
We propose a simple model to verify our assumption.
We train conditional Variational Auto-Encoders (cVAE) conditioned on the node semantic and geometric encoding $E$ to map $S$ to $P$. 
The encoder takes the $E$ and $V$ as input, outputting the mean $\mu$ and log-variance $\sigma$ for a Gaussian distribution, from which we sample a latent encoding $z$. $z$ is concatenated with $E$ and then fed to the decoder, getting the reconstructed $V'$. 
We adopted SMPL~\cite{SMPL} parameters as $V$. 
For a sample belonging to multiple nodes, we take the mean of their corresponding $E$ as the condition.
We train S2P on the 3D data of \textit{Pangea}, using KL divergence driving the predicted distribution to normal distribution and an L2 reconstruction loss of the SMPL parameters. 

The encoder and decoder in the cVAE are implemented as a 2-layer MLP.
The semantic and geometric encoders in P2S are \textit{frozen} during S2P.
The model is trained on \textit{Pangea} using an Adam optimizer for 100 epochs, with a batch size of 256.
The learning rate is warmed up from 5e-8 to 2e-6 for the initial 2 epochs and then decayed with a cosine scheduler.

\section{Datasets Details in Experiments}
\label{sec:detail_exper}
\textbf{HICO}~\cite{hico} is an \textit{image-level} benchmark for Human-Object Interaction (HOI) recognition. It has 38,116 and 9,658 images in the train and test sets and defines 600 HOIs composed of 117 verbs and 80 COCO objects~\cite{coco}. Each image has an image-level label which is the aggregation over all HOIs in an image without human boxes. We use mAP for multi-label classification.

\textbf{HAA}~\cite{haa500} is a video \textit{clip-level} human-centric atomic action dataset. It defined 500 actions and contains 10,000 video clips which are split into 8,000 training, 500 validating, and 1,500 testing clips. Each video clip has one single action label. The top-1 accuracy metric is utilized for multi-class classification.

\textbf{HMDB51~\cite{hmdb}} is a video \textit{clip-level} dataset consisting of 6,766 internet videos over 51 classes, and each video has from 20 to 1,000 frames. Each video clip has one single action label. We report the average top-1 accuracy on the standard three splits.

\textbf{Kinetics-400~\cite{kinetics400}} is a video \textit{clip-level} human-focused dataset that includes 240 K training clips and 20 K validation clips over 400 action classes. Each video lasts for about 10 seconds and contains one single label. We report the top-1 accuracy and top-5 accuracy on the official validation set as the convention.

\textbf{BABEL~\cite{BABEL:CVPR:2021}} is a large-scale 3D action dataset covering a wide range of human motions, including over 250 unique action classes. 
It is built upon AMASS~\cite{AMASS:ICCV:2019} by annotating the sequences with \textit{sequence-level} and \textit{frame-level} action classes, represented with the SMPL/SMPL-X body model~\cite{SMPL,SMPL-X}.
Over 43.5 hours of MoCap data is provided with 28,033 sequence labels and 63,353 frame labels and is categorized into one of 260 action classes. 
We follow the evaluation protocol of BABEL-120 under the dense label-only setting, containing a span of MoCap sequences belonging to 120 classes, where 13,320 sequences are divided into train (60\%), val (20\%), and test (20\%) sets. 
A motion-capture span of 5 seconds or less is given, and our model is required to predict the actions in it. 
Top-1 accuracy is reported. To show our ability in the long-tail classes, the Top-1-norm (the mean Top-1 across classes) is also reported.
We adopt PointNet++~\cite{PointNet++} trained on \textit{Pangea} as initialization and finetune it on BABEL.
Note that the BABEL-120 benchmark is based on motion sequences.
To adapt our model to the setting, we down-sample the original sequence from 60 FPS to 3 FPS, perform inference on all the down-sampled frames, and use \textit{mean pooling} to acquire the final score.

\textbf{HAA4D~\cite{tseng2022haa4d}} is an extension of HAA~\cite{haa500}. 
3,300 videos of 300 human atomic action classes from HAA are selected to construct a class-balanced and diverse dataset. 
Each video is annotated with globally aligned 4D human skeletons.
We follow the conventional action classification setting and data split~\cite{tseng2022haa4d}.
For classes containing 20 samples, the first 10 samples are adopted for training, and the rest are used for inference.
For classes containing 2 samples, the one with a bigger index is adopted for training, while the other one is adopted for inference.
We adopt PointNet++~\cite{PointNet++} pretrained on \textit{Pangea} as initialization and finetune it on HAA4D.
Since HAA4D only provides 4D skeletons, we fit the provided skeletons with SMPL~\cite{SMPL} and use the SMPL parameters for training and inference.
We perform inference on all the down-sampled frames and use \textit{mean pooling} to acquire the final score.

\section{Details of Image/Video Transfer Learning}
\label{sec:detail_analysis_transfer}
\subsection{Transfer Learning Stages}
P2S pretrained on \textit{Pangea} with node classification is a knowledgeable \textbf{backbone} and can be used in transfer learning.
There are three stages in transfer learning:
a) Training P2S on \textit{Pangea}, but with the val \& test sets of the downstream target dataset \textbf{excluded} following a strict transfer learning setting.
b) Finetuning P2S on the target dataset train set.
c) Training a small MLP  to transform $\mathcal{S}_{node}$ to $\mathcal{S}_{act}$, with the node prediction fixed.

\subsection{Training P2S}
\label{sec:train_p2s}
For the convenience of expression, we divide our \textit{Pangea} database in Suppl. Tab.~\ref{tab:dataset_statistics} into 4 splits: 
1) Willow Action~\cite{willow} $\sim$ HAKE~\cite{pastanet}: image datasets;
2) HMDB51~\cite{hmdb} $\sim$ Charades~\cite{charades}: video datasets with relatively small scale;
3) Charades-Ego~\cite{charades_ego} $\sim$ Kinetics~\cite{kinetics-700}: video datasets with relatively large scale;
4) HumanAct12~\cite{action2motion} $\sim$ HAA4D~\cite{tseng2022haa4d}: skeleton/MoCap datasets.

We select images from split 1\&2 to construct \textit{Pangea} test set to represent verb node semantics. The remaining images are used for training.
We first train a CLIP model with image-text pairs to get good physical representations, and then freeze the physical representations and train P2S.

To train physical representations, we use a CLIP pretrained ViT-B/32 image encoder to extract visual features with a resolution of 224. 
An AdamW~\cite{adamw} optimizer with a weight decay of 0.05 is used in training. 
We first use split 1\&2 data to train the model for 15 epochs with a batch size of 256 (split 3 is currently excluded to avoid the image domain gap). The learning rate is warmed up from 5e-7 to 1e-5 for the initial 2 epochs, then decayed with a cosine scheduler.
Then we use split 1\&2\&3 data to finetune the model for 50 epochs with a batch size of 256. 
The learning rate is warmed up from 5e-8 to 2e-6 for the initial 2 epochs, then decayed with a cosine scheduler. When training with split 1\&2\&3 data, a fixed number of samples from split 3 data are randomly sampled in each epoch for efficiency.

To train P2S, we freeze the physical representations and train the text encoders and the hyperbolic representations.
The model is trained for 5 epochs with a batch size of 64. The learning rate is warmed up from 5e-8 to 2e-6 for the initial 2 epochs, then decayed with a cosine scheduler.

Additionally, HMDB51 data is excluded from the training data to prevent data pollution because it has three train/test splits that intersect with each other. Also, for transfer learning on Kinetics-400, we use a new P2S model, where we exclude the data in Kinetics-700 but not in Kinetics-400 to prevent data pollution.

\subsection{Video Temporal Encoding}
For video benchmarks, we adopt lite implementations for temporal encoding and do not use video augmentation methods.
For the simplest temporal coding, we cut out fixed 8 frames for each video clip and \textit{average} logits of 8 frames as the clip logit. 
We compare several simple temporal coding methods on HAA~\cite{haa500} transfer learning.
\begin{itemize}
    \item Average prediction of frames. In training, supervision is applied to each frame. In testing, predictions of 8 frames are \textit{averaged} as the clip prediction. Our P2S achieves 71.40\% acc with this temporal encoding.
    \item Mean pooling. Frame-level visual features are first extracted, and the clip-level visual feature is obtained via simple mean pooling of frame-level ones. In training, supervision is applied to each clip. In inference, the clip prediction is directly outputted. With feature mean pooling, our P2S achieves 71.02\% acc.
    \item Temporal transformer. It is operated similarly to mean pooling, other than a temporal transformer inserted before the mean pooling of frame-level features. With the temporal transformer, our P2S achieves 71.47\% acc.
\end{itemize}
From the above results, we can find that with a more sophisticated model, the performance is higher too.
In future work, we believe a larger model with more computation power support will achieve more significant performance improvements with our \textit{Pangea}.
In this work, we report P2S results with \textit{average prediction of frames} temporal encoding for simplicity.
Even with a very simple temporal encoding, P2S performs comparably with some spatio-temporal (ST) methods.
P2S can also be used as a \textit{plug-and-play} method, we report the results of fusing P2S with SOTA video models.

\subsection{HICO}
With the pretrained P2S (Suppl.~Sec~\ref{sec:train_p2s}), we first finetune P2S on the HICO train set for 10 epochs,  with a batch size of 64. 
The learning rate is warmed up from 5e-7 to 1e-5 for the initial 2 epochs, then decayed with a cosine scheduler.
We then train the transformation from $\mathcal{S}_{node}$ to $\mathcal{S}_{act}$ with the node prediction fixed. The model is trained for 50 epochs,  with a batch size of 64 and a learning rate of 1e-4.

We find that HICO~\cite{hico} designed for human-object interaction (HOI) recognition (verb-object, \eg, \texttt{sit\_on-chair}) is more difficult than common action recognition (verb, \eg, \texttt{sitting}).
Moreover, most of \textit{Pangea} data are videos and thus have a larger domain gap with HICO.
Thus, compared with other video-based benchmarks, HICO~\cite{hico} benefits less from P2S pretraining.

\subsection{HAA} 
With the pretrained P2S (Suppl.~Sec~\ref{sec:train_p2s}), we conduct the transfer learning.
We finetune P2S on the HAA train set for 10 epochs,  with a batch size of 64. 
The learning rate is warmed up from 5e-7 to 1e-5 for the initial 2 epochs, then decayed with a cosine scheduler.
Then we train the transformation from $\mathcal{S}_{node}$ to $\mathcal{S}_{act}$ with the node prediction fixed. The model is trained for 40 epochs,  with a batch size of 64 and a learning rate of 2e-4.

For the experiments of integrating P2S with MLLM, we tried a SOTA MLLM: LLaMA Adapter V2~\cite{gao2023llama}.

When trained \textit{without} P2S, the backbone is finetuned on train set to output captions indicating the activity. The prompt is formulated as 
\begin{equation*}
\text{``Generate caption of this image".}
\end{equation*}
The model is required to answer:
\begin{equation*}
\text{``The image shows a person's activity: XXX."}
\end{equation*}(\eg ``The image shows a person's activity: shuffle\_dance.")
Then the top-1 accuracy is calculated by comparing the semantic distance between the output caption and ground-truth actions based on a CLIP~\cite{clip} ViT-B/32 pretrained text encoder. 

When trained \textit{with} P2S,
we formulate P2S prediction as a prompt and require the LLM to output the activity shown in the image. In detail, the prompt is formulated as 
\begin{gather*}
\text{``Some information related to the person's activity is: XXX.}\\\text{Describe the person's activity."}
\end{gather*}
The model is required to answer:
\begin{equation*}
\text{``The image shows a person's activity: XXX."}
\end{equation*}

To fuse the model w/ and w/o P2S, during inference, we ensemble the semantic distances between captions w/wo P2S and ground-truth action caption described above. For HICO\cite{hico}, as the model is required to give a list of HOIs containing one verb and one object each, we extract the HOIs from the generated caption, compare the semantic distances between predicted HOIs and GT HOIs, and calculate mAP following standard evaluation protocol. 

We prepare the data following the setting of stage 2/stage 1 for models w/wo P2S. The model is trained following the setting of stage 1 on both occasions.

\subsection{HMDB51} 
With the pretrained P2S (Suppl.~Sec~\ref{sec:train_p2s}), we first finetune P2S on HMDB51 train set for 10 epochs,  with a batch size of 64. 
The learning rate is warmed up from 5e-7 to 1e-5 for the initial 2 epochs, then decayed with a cosine scheduler.
Then we train the transformation from $\mathcal{S}_{node}$ to $\mathcal{S}_{act}$ with the node prediction fixed. The model is trained for 10 epochs,  with a batch size of 512. The learning rate is warmed up from 5e-7 to 1e-5 for the initial 2 epochs, then decayed with a cosine scheduler.

\subsection{Kinetics-400}
With the pretrained P2S (Suppl.~Sec~\ref{sec:train_p2s}), we conduct:
a) Finetuning P2S on Kinetics-400 train set for 15 epochs,  with a batch size of 192. 
The learning rate is warmed up from 1e-7 to 2e-6 for the initial 5 epochs, then decayed with a cosine scheduler.
b) Training the transformation from $\mathcal{S}_{node}$ to $\mathcal{S}_{act}$ with the node prediction fixed. The model is trained for 20 epochs,  with a batch size of 512. The learning rate is warmed up from 1e-7 to 2e-6 for the initial 5 epochs, then decayed with a cosine scheduler.

We find a decreased performance when pre-trained with CLIP-Pangea on Kinetics-400. This is possibly caused by the large data scale and complex action classes (400 total) of Kinetics-400 compared with other downstream datasets.

\section{Details of 3D Transfer Learning}\label{sec:detail_analysis_transfer_3d}
For 3D human point clouds, we use PointNet++~\cite{PointNet++} as the encoder. 
An AdamW~\cite{adamw} optimizer with a weight decay of 0.05 is used. 
The model is trained for 100 epochs with a batch size of 128. 
The learning rate is warmed up from 5e-8 to 2e-6 for the initial 2 epochs, then decayed with a cosine scheduler.
For P2S learning, we use 601 K 3D training human instances and test the model on \textit{Pangea} test set with 172 K 3D human instances.
About 75\% of the human instances are obtained from single-view reconstruction~\cite{romp,joo}.
We adopt GT 3D human for BABEL~\cite{BABEL:CVPR:2021} and use reconstructed 3D human for other datasets.

\subsection{BABEL}
To show the strength of P2S, we further conduct transfer learning on a large-scale 3D action dataset BABEL~\cite{BABEL:CVPR:2021}. We compare our method with the BABEL official baseline~\cite{BABEL:CVPR:2021}. 
We adopt 2s-AGCN as the baseline following BABEL~\cite{BABEL:CVPR:2021}, which utilizes temporal information. 
Besides, we use PointNet++ and CLIP as extra baselines.
Surprisingly, we find that the simple pipeline PointNet++ considerably outperforms its counterpart 2s-AGCN.
On one hand, we find that the baseline CLIP performs not well. The reason may be that, without enough 3D pretraining data, the image-based CLIP cannot adapt to the domain of BABEL well. It can be verified that CLIP-\textit{Pangea} performs much better and even outperforms PointNet++-\textit{Pangea} with the help of 3D pretraining samples from \textit{Pangea}.
On the other hand, PointNet++ performs much more robustly than CLIP as it is designed to encode the 3D point cloud information which suits this task better. However, they all perform worse than our P2S.
As shown, P2S without heavy temporal encoding outperforms all baselines. PointNet++-\textit{Pangea} and CLIP-\textit{Pangea} also show superiorities upon their original setting PointNet++ and CLIP thanks to the extensive knowledge from \textit{Pangea}.

\subsection{HAA4D}
Transfer learning is also conducted on the recently proposed 3D action dataset HAA4D~\cite{tseng2022haa4d}.
We compare our method with the HAA4D official baseline~\cite{tseng2022haa4d}. 
From the comparison of results, we draw a similar conclusion to the one on BABEL. 
As shown, competitive performance is achieved with the help of \textit{Pangea} pretraining for PointNet++ and CLIP. Meanwhile, the proposed methods such as disentangling, semantic, and geometric encoding help P2S further outperform all baselines and SOTA.
We also notice that the improvement on HAA4D of P2S upon the SOTA method SGN is relatively smaller.
We recognize the reason as two-fold.
First, HAA4D provides 3D keypoints as GT annotation, thus we have to fit the SMPL model to the keypoints for the SMPL parameters.
This results in noisy inputs for P2S.
Second, HAA4D tends to focus more on human atomic body motions.
The frames are therefore less discriminative, weakening the performance of our frame-level P2S on HAA4D.

\section{Additional Results of P2S and S2P}
\label{sec:add_result}
\subsection{Action Recognition with P2S} 
\begin{figure}[t]
    \centering
    \includegraphics[width=\linewidth]{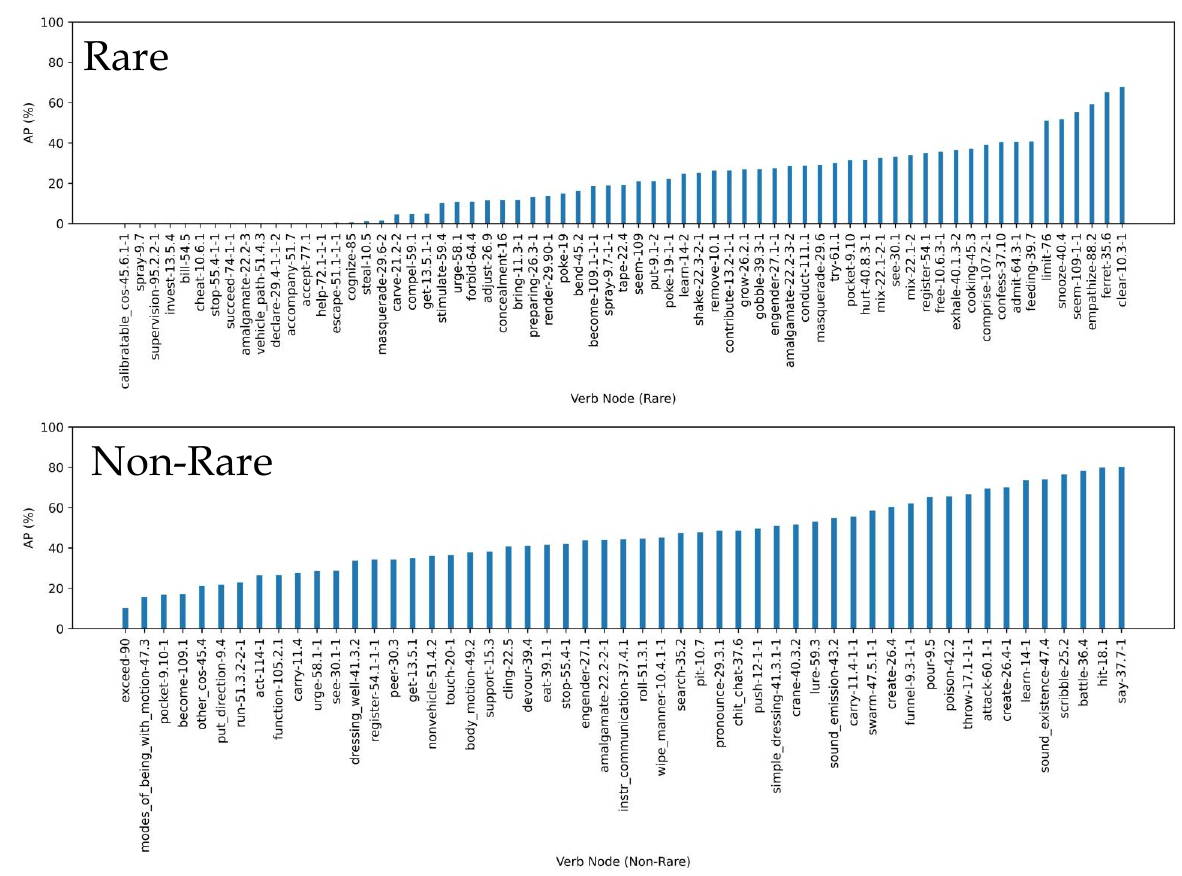}
    \caption{P2S performance on selected rare/non-rare verb nodes on Pangea benchmark. There are a total of 133 rare nodes and 157 non-rare nodes.}
    \label{fig:rare_nonrare}
\end{figure}

We list performance on selected rare/non-rare verb nodes on \textit{Pangea} benchmark in Suppl. Fig.~\ref{fig:rare_nonrare}. Our P2S achieves decent performance on both rare and non-rare verb nodes.

\begin{figure*}[t]
    \centering
    \includegraphics[width=0.85\linewidth]{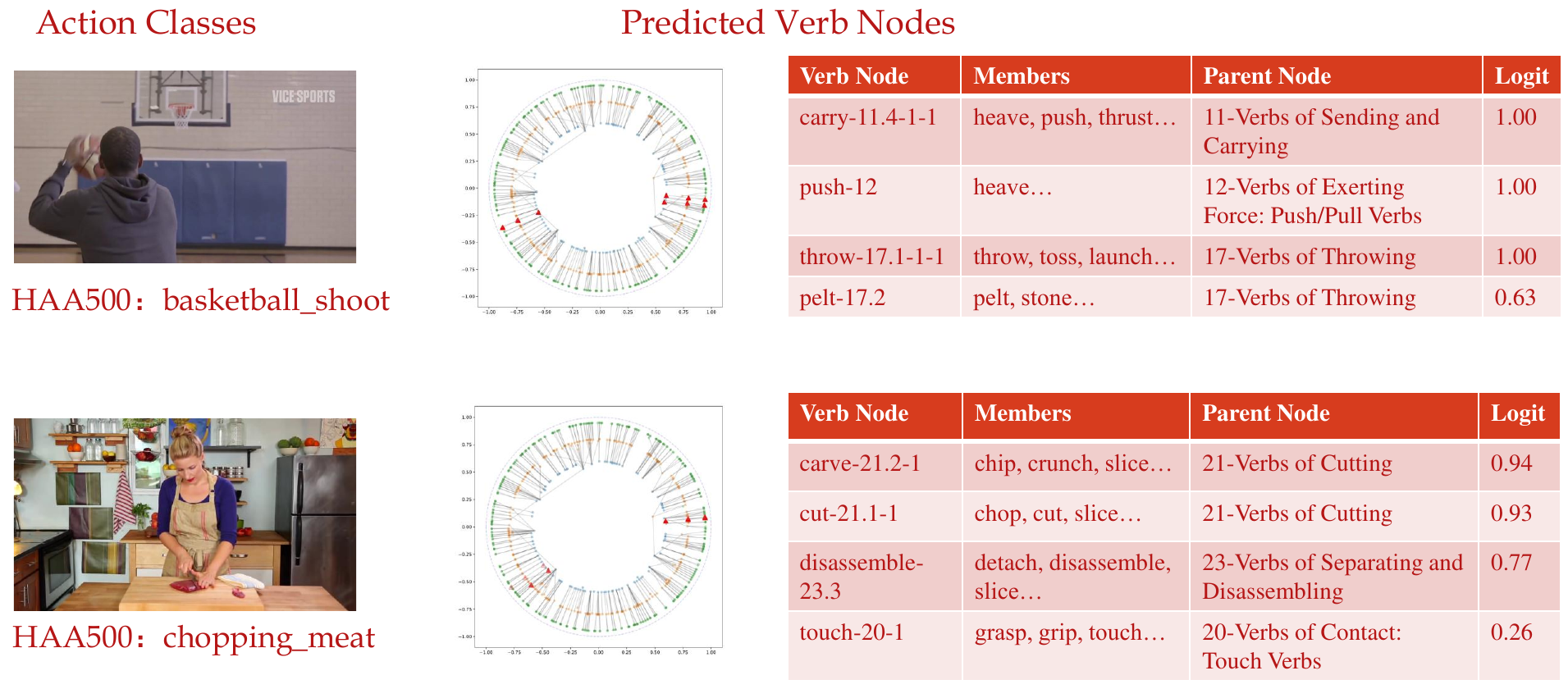}
    \caption{Example images and predicted verb node logits from the \textit{Pangea} test set. 
    }
    \label{fig:demo_figures}
\end{figure*}

\begin{figure*}[t]
    \centering
    \includegraphics[width=\linewidth]{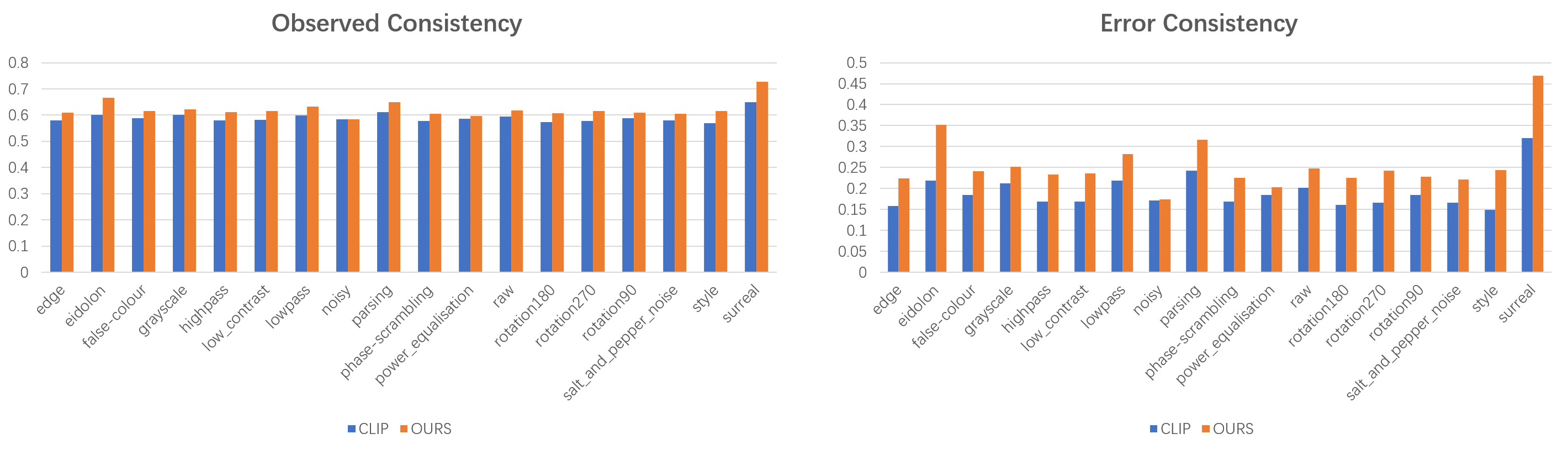}
    \caption{Consistency analysis upon CLIP and our method.}
    \label{fig:human_test_consistency}
\end{figure*}

Suppl.~Fig.~\ref{fig:demo_figures} further illustrates two examples of images and predicted verb node logits from the \textit{Pangea} test set. For each leaf node with high prediction, its verb members and parent node are shown.

We make further analysis and discussion of Pangea pre-training benefits as follows.
\textbf{a) 3D vs. 2D Benchmark:} 
P2S presents more evident performance improvement in 3D benchmarks mainly because:
\textit{1) Nature of tasks. }
Typically, the 3D benchmarks have smaller-scale train/test sets and simpler baselines than 2D image/video benchmarks. 
\textit{2) Smaller domain gap.} 
Various datasets from the Pangea database share SMPL parameters as 3D representations, whose domain gap is smaller than 2D image/frame pixels.
\textbf{b) Image vs. Video Benchmark: }
P2S performs better in image benchmark (HICO) as the baseline is a concise, image-based one, rather than a sophisticated video-based model.
\textbf{c) Variations within Video Benchmarks:}
P2S shows different benefits across video benchmarks since:
\textit{1) Size of pre-training data.}
For example, Kinetics-400 is a large-scale dataset. Pangea per-training data which is not from Kinetics-400 accounts for a relatively smaller proportion. 
\textit{2) Node samples distribution.}
Benchmarks are defined on their labels, thus mapped to different nodes. If its nodes have fewer samples in Pangea, the benchmark tends to \textit{benefit less} from P2S and perform worse. 
To roughly estimate the benefits from node samples distribution, for a benchmark $b$, we adopt an indicator $I_b=\sum_{i=1}^N cnt_{i}^{P} \cdot min(1, cnt_{i}^{b})$. Here, $N$: number of nodes, $cnt_{i}^{P}$/$cnt_{i}^{b}$: sample counts of node $i$ in Pangea/$b$, $min(1, cnt_{i}^{b})$: whether $b$ has samples for node $i$. 
For HAA/HMDB/Kinetics-400, $I_b$ is 93M/66M/92M. The gap between HAA (93M) and HMDB (66M) verifies this.

\subsection{P2S Consistency Analysis} 
To measure the robustness of models, we carry out a consistency test. We follow the setting of ~\cite{geirhos2021partial} and choose 100 head nodes from \textit{Pangea}. For each node, we chose 20 positive image samples and 20 negative samples. The negative samples are chosen from images of other nodes randomly. Each image has undergone \textbf{17 transformations}. 
The first 13 transformations are color-related transformations: grayscale, low contrast, noisy, salt and pepper noise, eidolon, false colourm, highpass, lowpass, phase scrambling, power equalization, rotating 90 degrees, rotating 180 degrees, and rotating 270 degrees.
The last 4 transformations are style changing, edge extracting, human parsing, and surreal. For the so-called surreal transformation, we grab a constructed 3D human mesh from one image and paste it into another background. 

\begin{figure*}[!h]
    \centering
    \includegraphics[width=0.85\textwidth]{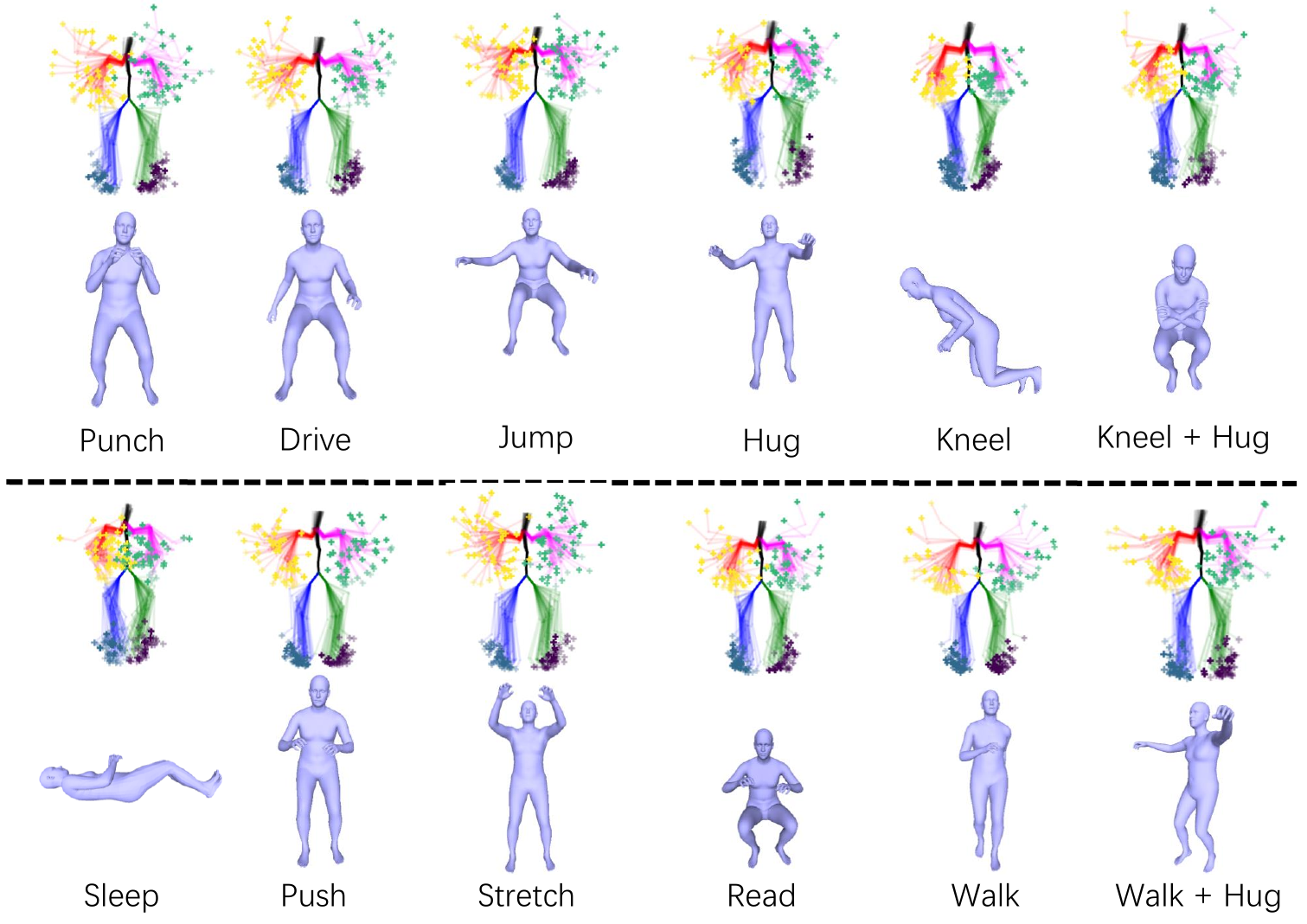}
    \caption{More S2P results.}
    \label{fig:add_s2p_vis}
\end{figure*}

Given the results of our method and the baseline CLIP,
we make an evaluation based on the metrics proposed in ~\cite{geirhos2021partial} and calculate the \textbf{observed consistency} and \textbf{error consistency}.
Observed consistency and error consistency are calculated concerning every node. For every node, the observed consistency is near or over 60\%, and the error consistency is between 20\% and 30\%. There are three transformations with \textit{striking high} consistency, namely human parsing, eidolon, and surreal. We believe that it is because these three transformations are too difficult.
Thus, we take the results of all nodes with the 3 weird high-consistency transformations deleted. The final results are shown in Suppl.~Fig.~\ref{fig:human_test_consistency}.

We can find that on both observed and error consistencies, our method P2S performs better than CLIP.
Thus, our method not only achieves better accuracy on recognition but also performs more robustly.

\begin{figure}[!h]
    \centering
    \includegraphics[width=0.5\textwidth]{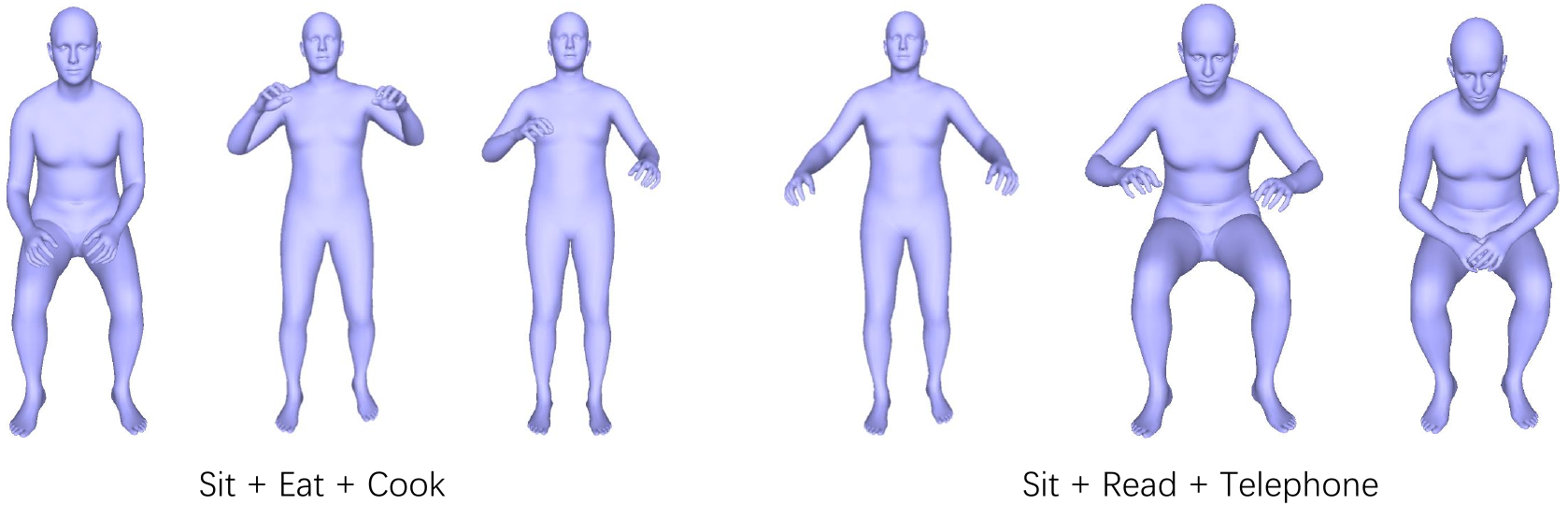}
    \caption{Failure cases of S2P.}
    \label{fig:s2p_failure}
\end{figure}

\subsection{3D Motion Generation with S2P}
We further visualize more results of S2P in Suppl.~Fig.~\ref{fig:add_s2p_vis}.

In detail, we align the samples by the pelvis joint, eliminate the root rotation along the z-axis to make the face orientation consistent, and draw skeletons for 100 samples of the \textit{same node} in the same figure to show the sample distribution.
As illustrated, S2P is capable of generating reasonable poses for various nodes.
And different nodes hold different geometric characteristics. 
For example, \texttt{ride} poses have elbows away from the spine; \texttt{sit} poses tend to have elbows near the spine; while there appears to exist more limb contraction for \texttt{kneel} and \texttt{sleep}.
Also, sample generation of node combination is also accessible.
By adding the condition \texttt{cellphone} upon \texttt{sit}, the wrist of the generated samples is restricted to distribute around the pelvis more.
Another interesting example is that adding \texttt{walk} upon \texttt{hug} amplifies the motion range.
We show rare combinations like \texttt{kneel} plus \texttt{hug}.
We also show some failure cases of our S2P in Suppl. Fig~\ref{fig:s2p_failure}.
As shown, when the node combination becomes more complicated, \eg, combining nodes with a larger semantic gap, our S2P could fail to generate accurate 3D actions. 
Here, we only use a simple cVAE to implement S2P. We believe more advanced models such as Transformer~\cite{posegpt} or Diffusion~\cite{zhang2022motiondiffuse} could generate more diverse and realistic 3D actions based on \textit{Pangea}. We leave this to future work.

\section{Additional Ablation Studies}
\label{sec:add_ablation}

\subsection{3D Representation in P2S}
To find the best feature extractor for 3D action data, we have tried different ways.
Specifically, we compared the performance of different representations of the 3D data: 
(i) SMPL\cite{SMPL} parameters, (ii) VPoser\cite{SMPL-X}, (iii) body key-points, and (iv) body point cloud. Note that our dataset only contains SMPL parameters and the other $3$ representations are all generated from the SMPL parameters.

For the first $3$ representations, we utilize two separate MLPs to encode and classify the 3D data. 
For the point cloud, we use the PointNet++\cite{PointNet++} as the 3D encoder, with an MLP as the classifier, which is referred to as Pointnet++ in the main text and Suppl. Tab.~\ref{tab:3d}.
Moreover, we also evaluate the CLIP-like classifier, where the cosine similarity between the encoded point cloud feature and the node semantic feature encoded by a text encoder is adopted as the final classification score.
This is referred to as CLIP in the main text and Suppl. Tab.~\ref{tab:3d}.
Suppl. Tab.~\ref{tab:3d} shows the results of different 3D representations on the Pangea Benchmark.
Specifically, in some instances of the Pangea dataset, ROMP~\cite{romp} fails to reconstruct 3D human bodies from the images. 
For these images, we eliminate these 3D data from the dataset during training and evaluation.

\begin{table}
\centering
\resizebox{0.4\textwidth}{!}{
\begin{tabular}{l c c c c }
    \hline
    Representation & Method  & Full & Non-Rare & Rare\\
    \hline
    \hline
    SMPL & MLP  & $8.32$  & $12.47$ & $3.42$ \\
    VPoser & MLP  & $7.81$  & $11.31$  & $2.55$ \\
    KeyPoints & MLP  & 5.45  & 8.44  & 1.92 \\
    Point Clouds & PointNet++ & $9.16$  & $12.76$  & $3.76$ \\
    Point Clouds & CLIP       & $\textbf{11.57}$  & $\textbf{16.12}$  & $\textbf{6.21}$ \\
    \hline
\end{tabular}}
\caption{Comparison of different 3D representations on Pangea benchmark.} 
\label{tab:3d}
\end{table}

Among these four representations, the point cloud achieves the best results. 
As for the method, we find that the performance of the model is further improved with a CLIP-like classifier.

We also evaluate the contribution of certain P2S components under the 3D only setting.
For example, without disentanglement, the performance degrades to \textbf{10.34} mAP, with a considerable performance decline of \textbf{2.51} mAP on the Rare set, proving the efficacy of our disentanglement strategy again.

\subsection{2D-3D Fusion in P2S}
We compare different 2D-3D fusion strategies in the P2S model. 
Note that since \textit{Pangea} contains data from different sources, some of which do not provide GT 3D human annotation, we adopt ROMP~\cite{romp} to generate pseudo 3D human annotations.
Suppl. Tab.~\ref{tab:fusion} shows the performance comparison of fusing the multi-modal data at different model stages. Early and middle fusion means that we fuse the extracted features of 2D and 3D at the early and middle layers of models respectively. For late fusion, we directly fuse the logits.
We can find that the late fusion which directly fuses the outputs of 2D and 3D models performs best.

\begin{table}[t]
    \centering
    \resizebox{0.7\linewidth}{!}{
    \begin{tabular}{l c c c }
        \hline
        Method  & Full & Non-Rare & Rare\\
        \hline
        \hline
        Early Fusion  & 37.08 & 48.05 & 24.12 \\
        Middle Fusion  & 36.30  & 47.37  & 23.23 \\
        Late Fusion  & \textbf{37.55}  & \textbf{48.84}  & \textbf{24.22} \\
        \hline
    \end{tabular}}
    \caption{Comparison of different multi-modal fusion strategies on \textit{Pangea} benchmark.} 
    \label{tab:fusion}
\end{table}

We also conduct a comparison between 2D only, 3D only, and 2D-3D fusion on \textit{Pangea} in Suppl. Tab.~\ref{tab:node_benchmark_modal}.
As shown, though P2S with 3D only is not very competitive by itself, they could still compensate for 2D only and bring considerable improvement.

\begin{table}[t]
    \centering
    \resizebox{0.7\linewidth}{!}{
    \begin{tabular}{l c c c}
        \hline
        Method  & Full & Non-Rare & Rare\\
        \hline
        \hline
        CLIP~\cite{clip}   & 28.25 & 37.87 & 16.90\\ 
        \hline
        P2S (2D)      & 34.46 & 45.15 & 21.84 \\ 
        P2S (3D)      & 11.57  & 16.12  & 6.21 \\
        P2S (2D+3D)   & \textbf{37.55}  & \textbf{48.84}  & \textbf{24.22} \\
        \hline
      \end{tabular}}
        \caption{Results of different modality utilization on \textit{Pangea}.} 
        \label{tab:node_benchmark_modal}
\end{table}

\subsection{Verb Node Encoding and Alignment in P2S}
\textbf{Entailment loss.}
Pangea faces a partial-label learning problem, where a few uncertain labels should have been annotated as True. 
The entailment loss, which enforces partial order relationships, adds more constraints than the classification loss. 
Thus, the entailment loss is only applied to positive samples to \textit{avoid over-constraints} on uncertain labels. 
We apply \textit{an additional ablation study}, where the entailment loss has an additional item for negative samples, following Eq.~32 from~\cite{desai2023hyperbolic}. 
We find a slight performance \textit{degradation} from \textit{34.25} mAP to \textit{33.93} mAP (with $\gamma=0.1$) on the Pangea test set.

\textbf{Semantic encoding flexibility.}
For semantic encoding, the input texts of the nodes are fixed, while the node embedding $E$ is not fixed with \textit{trainable} semantic and geometry encoding.
We use TextRank to sample key texts clarifying the node semantics better and taking the summarized text as the text encoder input, to include abundant information for semantic encoding.
To further explore semantic encoding flexibility, we conduct \textit{two additional experiments} compared with P2S (34.01 mAP) on the Pangea test set:
\textit{1) Augmentation via sampling node descriptions.} 
In training, a few randomly sampled sentences (up to 77 tokenized symbols) from the node descriptions are input.  
In inference, the fixed summarized text is input. 
The performance degrades to 32.43 mAP, possibly since the text bias from unrelated words in node descriptions: \eg, ``I put the book on the table'', ``book'' and ``table'' bring bias.
\textit{2) Use pretrained language vectors.} 
Each sentence is encoded via another pretrained CLIP text encoder, then the encoded features are fed into our text encoder as tokens. Thus, our text encoder receives various sentences as input.
The strategy performs comparably (33.96 mAP) with the original one. The possible reason is that there may be a trade-off between the encoded text length and learning difficulty. 
In the future, given more advanced LLMs to fully utilize the diverse semantic information of verb nodes in our database, we believe things may be different and will further stimulate the potential of our work.

\section{More Discussions} 
\label{sec:discussion}
In this section, we give some discussions about our system, some possible applications, and future studies based on our \textit{Pangea} and structured semantic space.

(1) Firstly, we discuss more possible \textbf{future applications} of our system as follows:

{\bf New Emergent and Very Rare Actions}.
Interestingly, we are creating new actions every day, \eg, new actions such as \texttt{play VR games, telesurgery} given the new inventions like VR player, telesurgery machine.  
These new emergent actions may have very limited visual and text data.
Given our structured semantic space, we can directly align new actions to their related verb nodes efficiently.
Then, we can easily find out the related/similar actions from the previous action database robustly instead of teaching machines a new action from scratch. 
The need for data collection would be largely reduced.
Moreover, it could alleviate the difficulty of incremental learning. 
Furthermore, sometimes it is very hard to collect data for very rare actions (\eg, \texttt{put out fire}), but we can get data easily from its parent, grandparent, or sibling nodes to help us gather its semantics. In inference, different levels of predictions also help because we can enforce their geometry relation consistency to get more robust results.

{\bf Customized Finetuning for Downstream Benchmarks}.
We can also customize the pretrain set for each downstream dataset. For example, for AVA, its classes are related to $n$ nodes in the tree. We can only collect the samples related to these $n$ nodes in our \textit{Pangea} and their closely related neighbors to build a customized and more powerful pretrain or train set for AVA. 

\textbf{Data Usage and Sharing}.
Given our \textit{Pangea}, it is easy to add new action data in pretraining or finetuning via the one-time verb node-class alignment. This provides a new solution for future applications to connect the data owners of different domains and fields. In the future, it is also promising to marry \textit{Pangea} and Federated learning to study data sharing and security.
Thus, we can build an action data platform to share and fully use data and evaluate the contributions of different data providers and annotators.

{\bf Training Considering Different Verb Tree Levels}.
Another possible application is that we can pretrain a model with high-level verb node labels only and then finetune it with finer-grained verb node labels. This follows the learning paradigm from abstract concepts to specific concepts. We leave this to future work.

{\bf Joint Learning of P2S and S2P}.
A promising application of our method is to jointly train P2S and S2P. For example, firstly train P2S and get the representative verb node features and then use it in S2P training. Secondly, we can generate new 3D human samples with S2P via distribution sampling. Next, these new 3D human samples can be input into P2S as pseudo samples. During the process, we can gradually add new data with labels to tune two models. 
This design may construct a loop to connect the bottom-up and top-down models and may show an interesting property.
It lays a foundation for a better understanding of the relationship between human geometry and behavioral semantics.

\textbf{Hyperbolic Embedding}.
Besides the geometry information encoding, the hyperbolic latent space also acts as an \textbf{interpretable indicator} to represent the action semantics and their change in images and videos, which is more than the performance gains. We think this would be vital for future general and interpretable action recognition studies.

{\bf Compositional Complexity}.
Human actions have compositional complexity at the human part level. 
On one hand, we can composite two actions such as \texttt{eat} and \texttt{walk} easily via human body parts control in 3D action generation.
On the other hand, this compositionality also brings challenges. Sometimes the label of a sample only reflects the action semantics carried by human parts, \eg, \texttt{hold} by hands, \texttt{kick} by feet. This phenomenon was studied by HAKE~\cite{hakev2,hakev1,pastanet} before. Given our structured action semantic space, we may be able to connect human body part states with our verb tree nodes to find out which nodes represent the part-level action semantics and which nodes carry the whole body semantics. 

(2) Next, we discuss the \textbf{design choices} of our system.

\textbf{3D Human}.
In our system, we use multi-modal inputs, \ie, 2D image/video and 3D human point cloud from SMPL mesh. Because we believe though 2D data carries abundant information about human actions, 3D human carries relatively more geometric information about human bodies. In our tests, we also find that they are complementary to each other. In the future, we believe that 3D action understanding will be a more and more important direction. Moreover, 3D action/motion generation has attracted more and more attention recently too.
Currently, we do not use the face and hand detection and reconstruction of 3D humans for simplicity. 
We can use a more advanced but also heavier whole body detection and 3D reconstruction model such as SMPLify-X~\cite{SMPL-X:2019}, to pursue better performance on face-hands related actions such as \texttt{eat, talk, grasp}, \etc. 
We leave this to future work.

{\bf Difference between CLIP-like Models and Ours}.
Action understanding has a long story but the semantic space is usually defined without guidance, \eg, selecting action classes according to the research interests or application requirements. Thus, different datasets cannot be directly used by other domains due to the action class setting divergence and semantic gap. This inhibits the development of general and open-action understanding. 
Recently, CLIP~\cite{clip} is proposed to utilize the flexible language prompt to encode the class labels, being able to bypass the class setting to achieve open-vocabulary training. But action semantics have their unique property overlooked by the intuitive visual-language alignment.
In detail, verbs usually have many senses under different contexts and scenes. Moreover, verb taxonomy is hierarchical, and different datasets usually adopt verbs in different granularities making the direct visual-language alignment difficult to capture the subtle semantics of actions. 
Directly using the label texts without any guidance is inefficient and makes it hard to scale for future large-scale applications. 
Recent works also find that CLIP-style works usually perform not as open as we thought since the confusion of competing text features~\cite{ren2022rethinking}. In our experiments, we also find that the ambiguity and complexity of action verbs and the obvious multi-label property of active persons hinder the effectiveness of CLIP a lot. 
In contrast, our structured semantic space design is explicit, well-designed to alleviate ambiguity, and relates the similar verbs thanks to the linguistic knowledge from VerbNet. Thus, our model performs much better than the vanilla CLIP design on large-scale action learning tasks while showing great generalization ability, openness, and extensibility~\cite{ren2022rethinking}.
Besides the unity and broad coverage, an extra benefit of our semantic space is that, though all the data would be placed in our verb tree, different users or researchers can only care about a part of the tree and do not need to process all the data of all the nodes while keeping the semantic structure knowledge. 

{\bf Weakly-Supervised Learning}.
In our \textit{Pangea}, due to the costly full annotation of the whole verb tree for all samples, we adopt a weakly supervised way to train the models. 
In the future, we can annotate more verb nodes for more action classes from existing datasets, supplement more node labels for the existing samples, or utilize the self-supervised learning method designed for the typical positive unlabeled setting (PU, only some of the positive samples have labels)~\cite{acharya2022positive} to further advance our weakly-supervised system. 

{\bf Long-tailed Distribution}.
Though we collect a lot of data in \textit{Pangea}, the distribution is still long-tailed due to the natural data distribution. 
However, in the future, the community can easily collect data for the rare nodes and train a more versatile model covering more nodes, and study more on how to generate better pseudo labels according to the language structure knowledge.

\end{document}

%% file: preamble.tex
\usepackage[dvipsnames]{xcolor}
